\RequirePackage{fix-cm}
\documentclass[twocolumn,twoside]{svjour3}          
\usepackage[switch]{lineno} 
\usepackage{graphicx}
\usepackage{comment}
\usepackage{amsmath,amssymb} 
\usepackage{color}
\usepackage[font=normalsize]{subfig}
\usepackage{blkarray}
\usepackage{threeparttable}
\usepackage{multirow}
\usepackage{makecell}
\usepackage{colortbl}
\usepackage{tikz}
\usepackage{fbox}
\usepackage{array}
\usepackage{booktabs}
\usepackage{wrapfig}
\usepackage{yhmath}
\usepackage{enumerate}
\usepackage{rotating}
\usepackage{diagbox}
\usepackage[authoryear]{natbib}

\usepackage{marginnote} 

\usepackage{url}
\usepackage[colorlinks,linkcolor=red,anchorcolor=blue,citecolor=blue]{hyperref}
\usepackage[linesnumbered,ruled,vlined]{algorithm2e}
\usepackage[misc]{ifsym}
\usepackage{circledsteps}

\usepackage{bbm}
\graphicspath{{fig/}}

\usepackage{mathtools}

\usepackage{verbatim}

\newcommand{\etal}{et al.}
\newcommand{\ie}{\textit{i.e.}}
\newcommand{\eg}{\textit{e.g.}}

 \journalname{International Journal of Computer Vision}
\begin{document}

\title{Deep Learning-based Image and Video Inpainting: A Survey}


\author{{Weize Quan} \textsuperscript{1,2}        \and
        {Jiaxi Chen} \textsuperscript{1,2} \and
        {Yanli Liu} \textsuperscript{3} \and
        {Dong-Ming Yan} \textsuperscript{1,2,~\Letter}  \and
        {Peter Wonka} \textsuperscript{4}
}


\institute{%
	\begin{itemize}
		\item[\textsuperscript{\Letter}] D.-M. Yan \\
		yandongming@gmail.com
		\at
		\item[\textsuperscript{1}] MAIS \& NLPR, Institute of Automation, Chinese Academy of Sciences, Beijing, China
		\item[\textsuperscript{2}] School of Artificial Intelligence, University of Chinese Academy of Sciences, Beijing, China
		\item[\textsuperscript{3}] College of Computer Science, Sichuan University, Chengdu, China
		\item[\textsuperscript{4}] Computer, Electrical and Mathematical Science and Engineering Division, King Abdullah University of Science and Technology, Thuwal, Saudi Arabia
	\end{itemize}
}

\date{Received: date / Accepted: date}

\maketitle

\begin{abstract}
Image and video inpainting is a classic problem in computer vision and computer graphics, aiming to fill in the plausible and realistic content in the missing areas of images and videos. With the advance of deep learning, this problem has achieved significant progress recently. The goal of this paper is to comprehensively review the deep learning-based methods for image and video inpainting. Specifically, we sort existing methods into different categories from the perspective of their high-level inpainting pipeline, present different deep learning architectures, including CNN, VAE, GAN, diffusion models, etc., and summarize techniques for module design. We review the training objectives and the common benchmark datasets. We present evaluation metrics for low-level pixel and high-level perceptional similarity, conduct a performance evaluation, and discuss the strengths and weaknesses of representative inpainting methods. We also discuss related real-world applications. Finally, we discuss open challenges and suggest potential future research directions.

\end{abstract}

\begin{figure}[t]
    \centering
    \includegraphics[width=1.\linewidth]{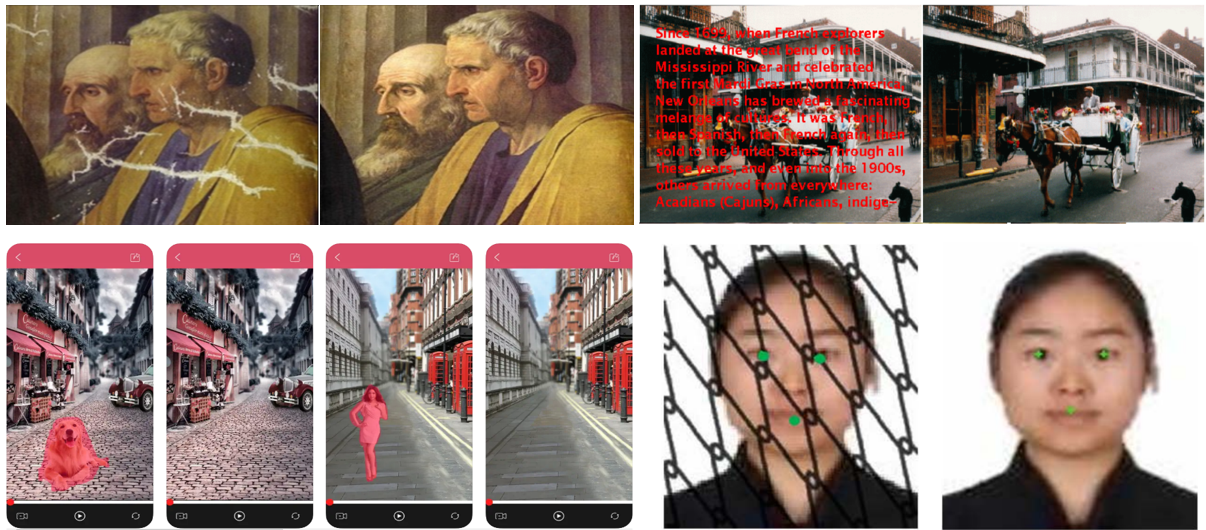}
    \caption{Application examples of inpainting techniques: photo restoration (top left: image from~\citep{bertalmio2000image}), text removal (top right: image from~\citep{bertalmio2000image}),  undesired target removal (bottom left: image from ~\citep{appstore}), and face verification (bottom right: image from ~\citep{Zhang2018DeMeshNet}).}
    \label{fig:image inpainting}
\end{figure}

\section{Introduction}
\label{sec:introduction}
\begin{figure}[t]
    \centering
    \includegraphics[width=\linewidth]{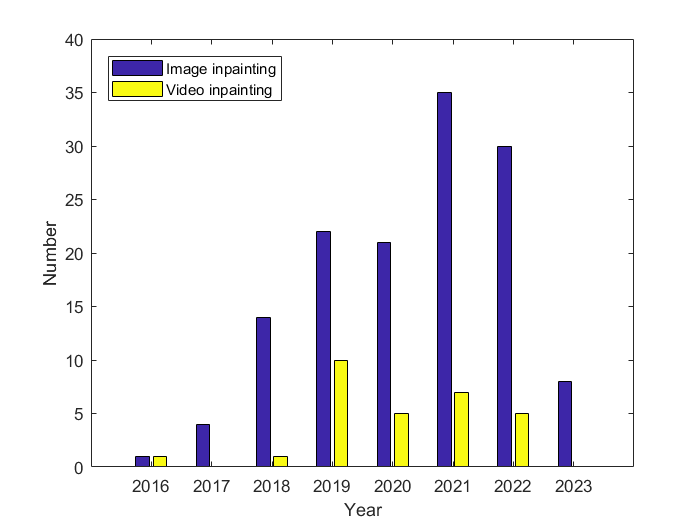}
    \caption{The rough number of papers on image and video inpainting per year.}
    \label{fig:paper_number}
\end{figure}

Image and video inpainting~\citep{Masnou1998level,bertalmio2000image} refers to the task of restoring missing/occluded regions of a digital image or video with plausible and natural content. Inpainting is an underconstrained problem with multiple plausible solutions, especially if there are large missing regions. Inpainting has many important applications in multiple fields, such as cultural relic restoration, virtual scene editing, digital forensics, and film and television production, etc. Fig.~\ref{fig:image inpainting} shows some important applications of inpainting techniques. Video is composed of multiple images exhibiting temporal coherence, therefore, video inpainting is closely related to image inpainting, where the former often learns from or extends the latter. For this reason, we simultaneously review image and video inpainting in this survey, and the number of papers is shown in Fig.~\ref{fig:paper_number}.


Early image inpainting methods mainly depend on low-level features of corrupted images, including PDE-based methods~\citep{bertalmio2000image,ballester2001filling, david2005vector} and patch-based methods~\citep{efros1999texture,barnes2009patchmatch,darabi2012image,huang2014image,jan2014high,guo2018patch}. PDE-based approaches usually propagate the information from the boundary to create a smooth inpainting. It is possible to propagate edge information, but it is hard to inpaint textures. Instead of only considering the boundary information, patch-based approaches recover the unknown regions by matching and duplicating similar patches of known regions. For smaller areas, patch-based methods can inpaint textures and also inpaint complete objects if similar objects are available in other image regions. However, these traditional methods have limited ability to generate new semantically plausible content, especially for large missing regions and missing regions that are not similar to other image regions. A comprehensive review on classical image inpainting methods is beyond our scope, and we refer readers to the surveys~\citep{guillemot2014image,jam2021a} for more details.

By contrast, deep learning holds the promise to inpaint large regions and also inpaint new plausible content that was learned from a larger set of images. In the beginning convolutional neural networks (CNNs) and generative adversarial networks (GANs) were the most popular choices in the inpainting literature. CNNs are a class of feed-forward neural networks that consist of convolutional, activation, and down-/up-sampling layers. They learn a highly non-linear mapping from the input image to the output image. GANs are a type of generative model consisting of a generator and a discriminator that estimates the data distribution through an adversarial process. Recently, more attention has been paid to the transformer architecture and generative diffusion models~\citep{sohl2015deep,DDPM}. Transformers are a prevalent network architecture based on parallel multi-head attention modules. Compared to the locality of CNNs, transformers have a better ability for contextual understanding. Diffusion probabilistic models are a type of latent variable model, which mainly contain the forward process, the reverse process, and the sampling procedure. Diffusion models learn to reverse a stochastic process (\ie, diffusion process) that progressively destroys data via adding noise. These deep learning-based image inpainting methods can achieve attractive results that surpass traditional methods in many aspects. From the perspective of the high-level inpainting pipeline, existing inpainting methods can be classified into three categories: a single-shot framework, a two-stage framework, and a progressive framework. Orthogonal to these main approaches, different technical methods can be observed in their realization, including mask-aware design, attention mechanisms, multi-scale aggregation, transform domain, deep prior guidance, multi-task learning, structure representations, loss functions, etc.

Compared with images, video data has an additional time dimension. Therefore, video inpainting not only fills in reasonable content in the missing regions for each frame but also aims to recover a temporally consistent solution. Because of this close relationship between image inpainting and video inpainting, many technical ideas used in image inpainting are often applied and extended to solve video inpainting tasks. Traditional video inpainting methods are usually based on patch sampling and synthesis~\citep{wexler2007space,granados2012background,Newson2014video,huang2016temporally}. These methods have limited ability to synthesize consistent content and capture complex motion and are often computationally expensive. To address these shortcomings, many deep learning-based methods have been proposed and achieved significant progress. There mainly exist four research directions: 3D CNN-based methods, shift-based methods, flow-guided methods, and attention-based methods. The core idea of these methods is to transfer information from neighboring frames to the target frame. 3D CNNs are the direct extension of 2D CNNs and work in an end-to-end manner. However, they often suffer from spatial misalignment and high computational cost. Shift-based methods can address these limitations to some extent, but within a limited temporal window only. Flow-guided approaches can produce higher resolution and temporally consistent results but are vulnerable to  
imperfect optical flow completion due to occlusion and complex motion. Attention-based methods fuse known information from short and long distances. Unfortunately, inaccurate attention score estimation often leads to blurry results. 

To our knowledge, there are several papers that review the deep learning-based inpainting works in the literature. \cite{elharrouss2020image} categorizes image inpainting methods into sequential-based, CNN-based, and GAN-based methods, and reviews related papers. To improve on their work, we also discuss common methodological approaches, loss functions, and evaluation metrics. We also add more discussion about further research directions and include newer work. \cite{jam2021a} reviews the traditional and deep learning-based image inpainting methods. However, they paid much attention to the traditional methods but have significantly fewer deep learning-based works compared to our survey. \cite{weng2022a} reviews some GAN-based image inpainting methods, but is generally shorter. Moreover, these existing surveys do not review the image and video inpainting simultaneously.

\section{Image Inpainting}
For the restoration of missing regions in an image, the results sometimes are not unique, especially for large missing areas. Consequently, there mainly exists two lines of research in the literature: (1) deterministic image inpainting and (2) stochastic image inpainting. Given a corrupted image, deterministic image inpainting methods only output an inpainted result while stochastic image inpainting methods can output multiple plausible results with a random sampling process. Inspired by multi-modal learning, some researchers have recently focused on text-guided image inpainting by providing additional information with text prompts.


\subsection{Deterministic Image Inpainting}
From the perspective of a high-level inpainting pipeline, existing works for deterministic image inpainting usually adopt three types of frameworks: single-shot, two-stage, and progressive. The single-shot framework usually adopts a generator network with a corrupted image as input and an inpainted image as output; The two-stage framework mainly consists of two generators, where the first generator achieves a rough result and then the second generator improves upon it; The progressive framework applies one or more generators to iteratively recover missing regions along the boundary.

\begin{figure}[t]
    \centering
    \includegraphics[width=\linewidth]{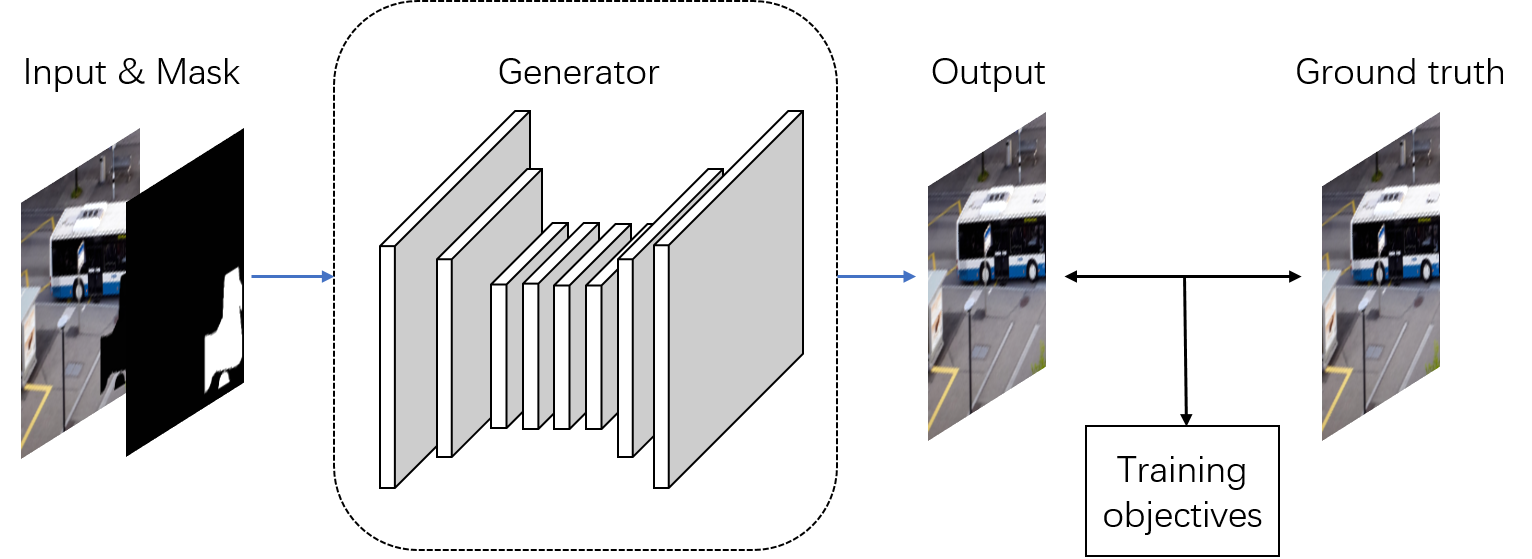}
    \caption{Representative pipeline of the single-shot inpainting framework. The generator takes as input the concatenation of a binary mask and a corrupted image and outputs the completed image. Training objectives are used for training the generator. }
    \label{fig:single_shot}
\end{figure}

\subsubsection{Single-shot framework} 
Many existing inpainting methods adopt a single-shot framework, as shown in Fig.~\ref{fig:single_shot}. It essentially learns a mapping from a corrupted image to a complete image. The framework usually consists of generators and corresponding training objectives. 

\textbf{Generators.}
To improve the inpainting ability of the generator, there exist several lines of research: mask-aware design, attention mechanism, multi-scale aggregation, transform domain, encoder-decoder connection, and deep prior guidance.

(1) Mask-aware design. \\
The missing regions (indicated with a binary mask) have different shapes and convolutional operations overlapping with these missing regions may be the source of visual artifacts. Therefore, some researchers proposed mask-aware solutions for classical convolutional operation and normalization. Inspired by the inherent spatially varying property of image inpainting, \cite{ren2015shepard} designed a Shepard interpolation layer where the feature map and mask both conduct the same convolution operation. Its output is the fraction of feature convolution and mask convolution results. Mask convolution can simultaneously update the mask. To better handle various irregular holes and evolve the hole during mask updating, \cite{liu2018image} proposed a mask-guided convolution operation, \ie, partial convolution, which distinguishes between the valid region and hole in a convolutional window. \cite{xie2019image} proposed trainable bidirectional attention maps to extend the partial convolution~\citep{liu2018image}, which can adaptively learn the feature re-normalization and mask-updating. 

Different from the feature normalization considered by previous methods, \cite{yu2020region} focused on the mean and variance shift-related normalization and introduced a spatial region-wise normalization into the inpainting network. \cite{wang2020vcnet} designed a visual consistency network for blind image inpainting. They first predicted the damaged regions yielding a mask, and then applied an inpainting network with the proposed probabilistic context normalization, which transfers the mean and variance from known features
to unknown parts building on different layers. Inspired by filling holes with pixel priorities~\citep{criminisi2004region,zhang2019gain}, \cite{wang2021parallel} used a structure priority (in low-resolution features) and a texture priority (in high-resolution features) in partial convolution~\citep{liu2018image}. \cite{Wang2021dynamic} proposed a dynamic selection network to utilize the valid pixels better. Specifically, they designed a validness migratable convolution to dynamically sample the convolutional locations, and a regional composite normalization module to adaptively composite batch, instance, and layer normalization on mask-based selective feature maps. \cite{zhu2021image} learned to derive the convolutional kernel from the mask for each convolutional window and proposed a point-wise normalization that produces the mask-aware scale and bias for batch normalization.

(2) Attention mechanism.\\ 
Attention is a prevalent tool to model correlation in the field of natural language processing~\cite{Vaswani2017Attention} and computer vision~\citep{Wang2018Nonlocal,fu2019dual}. Attention is better at accessing features of distant spatial locations than convolution. In the literature, \cite{yu2018generative} was the first to introduce a contextual attention mechanism to image inpainting. This pioneering work inspired many following works. To enhance both visual and semantic coherence, \cite{zeng2019learning} proposed a pyramid-context encoder network with an attention transfer method, where the attention score computed in a high-level feature is used for low-level feature updating. Instead of using one fixed patch size for attention computation, \cite{wang2019multi} proposed a multi-scale contextual attention model with two different patch sizes followed by a channel attention block~\citep{hu2018squeeze}. \cite{wang2020multi} introduced a multistage attention module that performs large patch swapping in the first stage and small patch swapping in the next stage. \cite{qin2021multi} combined spatial-channel attention~\citep{Chen2017SCACNN} and a spatial pyramid structure to construct a multi-scale attention unit (MSAU). This unit separately conducts spatial attention on four feature maps obtained by different dilation convolutions and then applies augmented channel attention on concatenated attentive features. \cite{zhang2022wnet} proposed a structure and texture interaction network for image inpainting. They designed a texture spatial attention module to recover texture details with robust attention scores guided by coarse structures and introduced a structure channel excitation module to recalibrate structures according to the difference between coarse and refined structures.

In addition, some recent works proposed image inpainting networks based on vision transformers~\citep{dosovitskiy2021an}. \cite{deng2021learning} proposed a contextual transformer network to complete the corrupted images. Their network mainly depends on the multi-scale multi-sub-head attention, which is extended from the original multi-head attention proposed by~\citep{Vaswani2017Attention}. \cite{cao2022learning} incorporated rich prior information from the ViT-based masked autoencoder (MAE)~\citep{He2022Masked} into image inpainting. Specifically, the pre-trained MAE model provides the features prior to the encoder of the inpainting network and the attention prior to making the long-distance relationship modeling easier. Instead of using shallow projections or large receptive field convolutions to sequence the incomplete image, \cite{zheng2022bridging} designed a restrictive CNN head with a small and non-overlapping receptive field as token representation. \cite{deng2022hourglass} modified multi-head self-attention by inserting a Laplace distance prior, which computes the similarity considering the features and their locations simultaneously.

(3) Multi-scale aggregation.\\
In the literature on image processing, multi-scale aggregation is a common method to fuse information from different resolutions. \cite{wang2018image} designed a generative multi-column inpainting network, consisting of three convolution branches with different filter kernel sizes, to fuse multi-scale feature representations. To create a smooth transition between the inpainted regions with existing content, \cite{Hong2019deep} proposed a deep fusion network with multiple fusion modules and reconstruction loss applied on multi-scale layers. The fusion module merged predicted content with the input image via a learnable alpha composition. \cite{hui2020image} proposed a dense multi-scale fusion module, which fuses hierarchical features obtained by multiple convolutional branches with different dilation rates. \cite{zheng2021gcmnet} designed a progressive multi-scale fusion module to extract multi-scale features in parallel and progressively fuse these features, yielding more representative local features. Inspired by the high-resolution network (HRNet) for visual recognition~\citep{sun2019deep,wan2021high}, \cite{wang2021parallel} introduced a parallel multi-resolution fusion network for image inpainting. This network can simultaneously conduct inpainting in multiple resolutions with mask-aware and attention-guided representation fusion methods. \cite{phutke2021diverse} also followed a multi-path design, where they introduce four concurrent branches with different resolutions in the encoder. A residual module with diverse receptive fields is designed as the building block of the encoder. \cite{cao2021learn} proposed a multi-scale sketch tensor network for man-made scene inpainting. This network reconstructs different types of structures by adding constraints on predicted lines, edges, and coarse images with different scales. Different from the mask-blind processing~\citep{li2020deepgin,qin2021multi} of multi-scale features produced by convolution with different dilation rates, \cite{zeng2022aggregated} carefully designed a gated residual connection, which considers the difference between holes and valid regions. They also proposed a soft mask-guided PatchGAN, where the discriminator is trained to predict the soft mask obtained by Gaussian filtering.

(4) Transform domain.\\
Instead of conducting image inpainting in the spatial domain, some existing works designed inpainting frameworks in a transformed domain via the DWT (discrete wavelet transform)~\citep{dwt} and the FFT (fast Fourier transform). \cite{wang2020image} recast the image inpainting problem as predicting low-frequency semantic structures and high-frequency texture details. Specifically, they decomposed the corrupted image into different frequency components via the Haar wavelet transform~\citep{haar}, designed a multi-frequency probabilistic inference model to predict the frequency content in missing regions, and inversely transformed back to image space. \cite{yu2021wavefill} adopted a similar inpainting pipeline. For the multi-frequency completion, they proposed a frequency region attentive normalization module to align and fuse the features with different frequencies and applied two discriminators to two high-frequency streams. \cite{li2021detail} extracted high-frequency subbands as the texture and introduced a DWT loss to constrain the fidelity of low- and high-frequency subbands. LaMa~\citep{lama} combined the residual design~\citep{he2016deep} and fast Fourier convolution~\citep{ffc} to construct a fast Fourier convolution residual block, which is integrated into the encoder-decoder network to handle large mask inpainting. \cite{lu2022GLaMa} further improved LaMa by introducing various types of masks and adding the focal frequency loss~\citep{jiang2021focal} to constrain the spectrum of the images.

(5) Encoder-decoder connection. \\
Some works modify the basic encoder-decoder architecture by introducing carefully designed feature connections. Shift-Net~\citep{yan2018shift} modified the U-Net architecture by introducing a specific shift-connection layer, which shifts the encoder features of the valid region to the missing regions with a guidance loss. \cite{Dolhansky2018eye} introduced an eye inpainting network that merges the identifying information of the reference image encoding as a code. \cite{shen2019single} designed a densely connected generative network for semantic image inpainting. They combined four symmetric U-Nets with dense skip connections. \cite{liu2020rethinking} introduced a mutual encoder-decoder CNN, fusing the texture and structure features (from the shallow and deep layers of the encoder), to jointly restore the structure and texture with feature equalization. Similarly, \cite{guo2021image} designed a two-stream image inpainting network, which combines a structure-constrained texture synthesis submodel and a texture-guided structure reconstruction submodel. In addition, they introduced a bi-directional gated feature fusion module and a contextual feature aggregation module to fuse and refine the resulting images. \cite{xin2022generative} inserted generative memory into the classical encoder-decoder network to jointly exploit the high-level semantic reasoning and the pixel-level content reasoning. Based on~\citep{liu2020rethinking}, \cite{liu2022reference} inferred the texture and structure with a content-consistent reference image through a feature alignment module.

(6) Deep prior guidance. \\
To enhance the performance of the inpainting generator, some works have explored the deep prior from a single image or a large image database. \cite{Lempitsky2018deep} utilized a randomly-initialized generator network as the prior to completing the corrupted image by only reconstructing the known regions. \cite{gu2020image} proposed mGANprior by incorporating a pre-trained GAN as prior for image inpainting. Specifically, this method reconstructs the incorrupt regions while filling in the missing areas by adaptively merging multiple generative feature maps from different latent codes. \cite{pSp2021} developed a pixel2style2pixel (pSp) framework for image inpainting. They introduced an encoder consisting of a feature pyramid and multiple mapping networks to encode the damaged image into extended latent space $\mathcal{W}+$ (18 512-dimensional style vectors), which is the extension of latent space $\mathcal{W}$~\citep{Karras2019a}, and reused a pre-trained StyleGAN generator as priors to achieve the complete image. To handle the large missing regions and complex semantics, \cite{wang2022dual} designed a dual-path image inpainting framework with GAN inversion~\citep{xia2022gan}. Given a corrupted image, the inversion path infers the close latent code and extracts the corresponding multi-layer features from the trained GAN model, and the feed-forward path fills the missing regions by merging the above semantic priors with a deformable fusion module. To guarantee the invariance of the valid area in the corrupted and completed images, \cite{yu2022high} modified the GAN inversion pipeline~\citep{pSp2021} by designing the mapping network with a pre-modulation module and introducing $\mathcal{F} \& \mathcal{W}+$ latent space, where $\mathcal{F}$ are the feature maps of the corrupted image.

\textbf{Training objectives.}
The training objective is a very important component of deep learning-based image inpainting methods. Pixel-wise reconstruction loss, perceptual loss~\citep{johnson2016perceptual}, style loss~\citep{Gatys2016image}, and adversarial loss~\citep{goodfellow2014generative} are the prevalent training objectives. The adversarial loss is obtained by a discriminator network. \cite{pathak2016context} and \cite{li2019context} adopted the discriminator (stacked convolution and down-sampling) from DCGAN~\citep{radford2015unsupervised}. Considering \cite{pathak2016context}'s method struggles to maintain local consistency with the surrounding regions, \cite{iizuka2017globally} proposed local and global discriminators, which generate more realistic contents. \cite{yu2018generative} proposed a patch-based discriminator, which can be regarded as the generalized version of local and global discriminators~\citep{iizuka2017globally}. This patch-based discriminator is subsequently used in many following works. \cite{liu2021fine} designed two discriminators with small- and large-scale receptive fields to guide the inpainting network for fine-grained image detail generation.  

Besides, researchers have also introduced some carefully designed losses. \cite{li2017generative} introduced a semantic parsing loss for face completion. \cite{Yeh2017semantic} proposed context and prior losses to search the closest encoding in the latent image manifold for inferring the missing content. \cite{vo2018struc} proposed a structural reconstruction loss, which is the combination of reconstruction errors in pixel and feature space. 
For explicitly exploring the structural and textural coherence between filled contents and their surrounding contexts, \cite{li2019generative} utilized the local intrinsic dimensionality~\citep{lid1,lid2} in the image- and patch-level to measure and constrain the alignment between data submanifolds of inpainted contents and those of the valid pixels. To stabilize the training process of face inpainting, \ie, weakening gradient vanishing and model collapse, \cite{han2021face} trained the generator via neuro-evolution and optimized the generator's parameters by mutation and crossover. 


Some researchers introduced additional training objectives via multi-task learning. \cite{liao2018face} presented a novel collaborative framework by training a generator simultaneously on multiple tasks, \ie, face completion, landmark detection, and semantic parsing. To enhance the inpainting capability of the network for image structure, \cite{jie2020inpainting} designed a structure restoration branch in the decoder and explicitly inserted the structure features into the primary inpainting process. Appropriate semantic guidance is a suitable tool for image inpainting~\citep{song2018spg}, inspired by this, \cite{liao2020guidance,liao2021uncertainty,liao2021image} proposed a unified framework to jointly predict the segmentation maps and recover the corrupted images. Specifically, \cite{liao2020guidance,liao2021uncertainty} designed a semantic guidance and evaluation network that iteratively updates and evaluates a semantic map and infers the missing contents in multiple scales. However, this method may create implausible textures and blurry boundaries, especially on mixed semantic regions. To solve this problem, \cite{liao2021image} devised a semantic-wise attention propagation module to apply the attention operation on the same semantic regions. They also introduced two coherence losses to constrain the consistency between the semantic map and the structure and texture of the inpainted image. \cite{zhang2020pixel} studied how to improve the visual quality of inpainted images and proposed a pixel-wise dense detector for image inpainting. This detection-based framework can localize the artifacts of completed images, and the corresponding position information is combined with the reconstruction loss to better guide the training of the inpainting network. \cite{zhang2021context} introduced the semantic prior estimation as a pretext task with a pre-trained multi-label classification model, and then utilized the learned semantic priors to guide the inpainting process through a spatially-adaptive normalization module~\citep{spade}. \cite{yu2022unbiased} jointly solved image reconstruction, semantic segmentation, and edge texture generation. Each branch is implemented with a transformer network, and a multi-scale spatial-aware attention block is developed to guide the main image inpainting branch from the other two branches. Similar to~\citep{zhang2020pixel}, \cite{zhang2022perceptual} first localized the perceptual artifacts from the completed image, and then used this information to guide the iterative refinement process. They also manually annotated an inpainting artifact dataset.

\begin{figure*}[t]
    \centering
    \subfloat[Coarse-to-fine]{
    \includegraphics[width=0.9\textwidth]{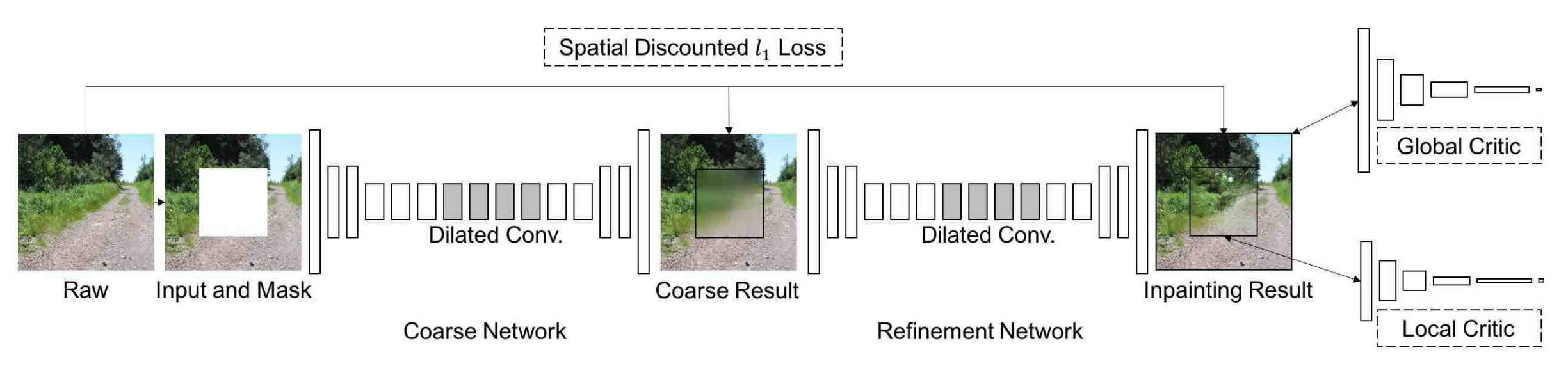}
    } \\
    \subfloat[Structure-then-texture]{
    \includegraphics[width=0.9\textwidth]{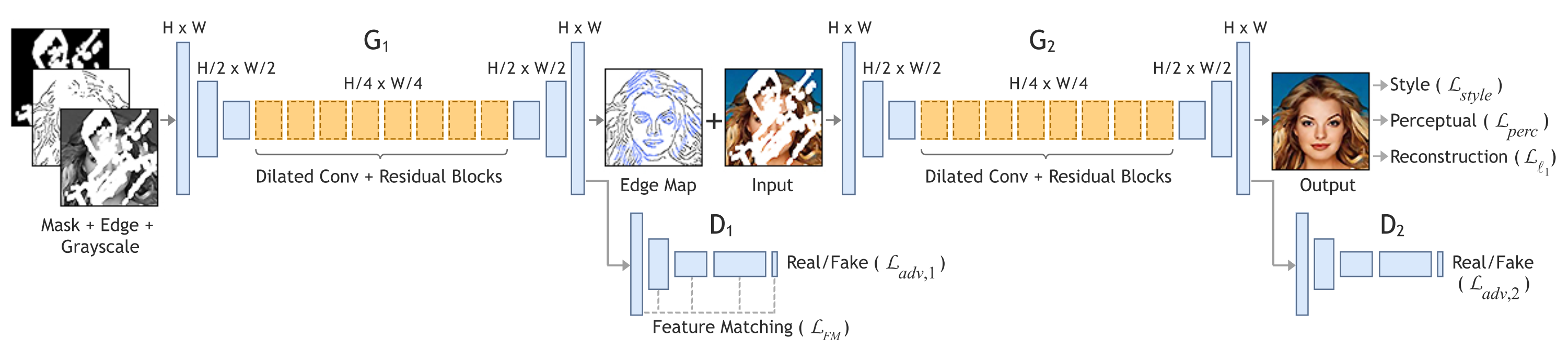}
    }
    \caption{Two types of the two-stage inpainting framework: (a) coarse-to-fine~\citep{yu2018generative} where the first network predicts an initial coarse result and the second network predicts a refined result; (b) structure-then-texture~\citep{nazeri2019edgeconnect} where the first network predicts a structure map and the second network predicts a complete image. An apparent difference between these two types is that the structure-then-texture methods explicitly predict the structure map in the first stage.}
    \label{fig:two_stage}
\end{figure*}

\subsubsection{Two-stage framework}

\textbf{Coarse-to-fine methods.} This kind of method first applies a generator to fill the holes with coarse contents, and then refine them via the second generator, as shown in Fig.~\ref{fig:two_stage}(a). \cite{yu2018generative} modified the generative inpainting framework with cascaded coarse and refinement networks. In the refinement stage, they designed a contextual attention module modeling the long-term correlation to facilitate the inpainting process. 

Many later works refined different aspects of this classical coarse-to-fine framework. Inspired by mask-aware convolution~\citep{liu2018image} for irregular holes, \cite{yu2019free} improved the previous network~\citep{yu2018generative} by introducing gated convolution that adaptively perceives the mask location. In the coarse stage, \cite{ma2019coarse} proposed region-wise convolutions and a non-local operation to process the discrepancy and correlation between intact and damaged areas. PEPSI \cite{sagong2019pepsi} modified the two-stage feature encoding processes in \citep{yu2018generative} by sharing the encoding network and organizing the coarse and fine inpainting network in a parallel manner. PEPSI can enhance the inpainting capability while reducing the number of convolution operations and computational resources. To further reduce the network parameters, \cite{shin2021pepsi} extended PEPSI by replacing the original dilated convolutional layers~\citep{dilated2016} with a so-called rate-adaptive version, which shares the weights for each layer but produces dynamic features via dilation rates-related scaling and shifting operations. The contextual attention proposed by~\citep{yu2018generative} has a limited ability to model the relationships between patches inside the holes, therefore, \cite{liu2019coherent} introduced a coherent semantic attention layer, which can enhance the semantic relevance and feature continuity in the attention computation of hole regions. In~\citep{yu2018generative}, several dilated convolutions are applied to enlarge the receptive field. \cite{li2020deepgin} replaced the dilated convolution with a spatial pyramid dilation ResNet block with eight different dilation rates to extract multi-scale features. \cite{Navasardyan2020image} designed a patch-based onion convolution mechanism to continuously propagate information from known regions to the missing ones. This convolution mechanism can capture long-range pixel dependencies and achieve high efficiency and low latency. ~\cite{wadhwa2021hyper} proposed a hypergraph convolution with a trainable incidence matrix to generate globally semantic completed images and replaced the regular convolutions with gated convolution in the discriminator to enhance the local consistency of inpainted images.

Due to the computational overhead and the lack of supervision for the contextual attention in~\citep{yu2018generative}, \cite{zeng2021cr} removed this attention block and learned its patch-borrowing behavior with a so-called contextual reconstruction loss. Based on the insight that recovering different types of missing areas need a different scope of neighboring areas, \cite{quan2022image} designed a local and global refinement network with small and large receptive fields, which can be directly applied to the end of existing networks to further enhance their inpainting capability. \cite{kim2022zoom} developed a coarse-super-resolution-refine pipeline, where they add a super-resolution network to reconstruct finer details after the coarse network and introduce a progressive learning mechanism to repair larger holes.

Some works adopt a coarse-to-fine framework to obtain high-resolution inpainting. \cite{yang2017high} designed a two-stage inpainting framework consisting of a content network and a texture network. The former predicts the holistic content in the low resolution ($128 \times 128$) and the latter iteratively optimizes the texture details of missing regions from low to high resolution ($512 \times 512$). \cite{song2018contextual} developed an image-to-feature network to infer coarse results, and then designed a patch-swap method to refine the coarse features. The swapped feature map is translated to a complete image via a Feature2Image network. In addition, this framework can be directly used for high-resolution inpainting by upsampling the complete image as the input of refine stage with a multi-scale inference. \cite{yi2020contextual} proposed a contextual residual aggregation mechanism for ultra high-resolution image inpainting (up to 8K). Specifically, a low-resolution inpainting result was first predicted via a two-stage coarse-to-fine network and then the high-resolution result was generated by adding the large blurry image with the aggregated residuals, which are obtained by aggregating weighted high-frequency residuals from contextual patches. \cite{zhang2022inpainting} focused on image inpainting for 4K or more resolution. They first fill the hole via LaMa~\citep{lama}, predict depth, structure, and segmentation map from the initially completed image, then generate multiple candidates with a multiply-guided PatchMatch~\citep{barnes2009patchmatch}, and finally choose a good output using the proposed auto-curation network. To complete the high-resolution image with limited resources, these methods first predicted the coarse content at the low-resolution level and then refine the texture details at the high-resolution level (sometimes with multi-scale inferences).

Other works also follow the basic coarse-to-fine strategy, but they are clearly different from the framework proposed by~\citep{yu2018generative}. After obtaining the coarse result with an initial prediction network, \cite{yang2019fine} applied a super-resolution network as the refinement stage to produce high-frequency details. \cite{Roy2021image} predicted the coarse results in the frequency domain by learning the mapping of the DFT of the corrupted image and its ground truth. Based on the insight that patch-based methods~\citep{barnes2009patchmatch,he2012statistics} fill the missing regions with high-quality texture details, \cite{xu2021texture} proposed a texture memory-augmented patch synthesis network with a patch distribution loss after the coarse inpainting network.

\textbf{Structure-then-texture methods.} Structure and texture are two important components of the image, therefore, some works decompose the image inpainting as the structure inference and the texture restoration, as shown in Fig.~\ref{fig:two_stage}(b). \cite{sun2018natural} designed a two-stage head inpainting obfuscation network. The first stage generates facial landmarks and the second stage recovers the head image guided by the landmarks. \cite{song2019geometry} first estimated the facial geometry including landmark heatmaps and parsing maps, and then concatenated these results with a corrupted face image as the input of the complete network to recover face images and disentangle masks. \cite{liao2018edge} and \cite{nazeri2019edgeconnect} both proposed an edge-guided image inpainting method, which first estimates the edge map for the missing regions, and then utilizes this edge map prior to predicting the texture details. Similarly, \cite{xiong2019foreground} explicitly disentangled the image inpainting problem into two sub-tasks of foreground contour prediction and content completion. To improve the structural guidance of coarse edge maps, \cite{ren2019structureflow} introduced another representation of the structure, \ie, the edge-preserving smoothing via filtering operation. Based on the structure reconstruction of the first network, they inpainted missing regions using appearance flow. \cite{shao2020generative} combined the edge map and color aware map as the representation of the structure, where the former is captured via the Canny operator~\citep{canny1986computational} and the latter is obtained through Gaussian blur with a large kernel. For the specific Manga inpainting, \cite{xie2021seamless} first completed a semantic structure map, including the structural lines and the ScreenVAE map (a point-wise representation of screentones)~\citep{xie2020manga}, using a semantic inpainting network. Then, the completed semantic map is used for guiding the appearance synthesis. \cite{wang2021image} designed an external-internal learning inpainting framework. It first reconstructs the structures in the monochromatic space using the knowledge externally learned from large datasets. Based on internal learning, then, it applies a multi-stage network to recover the color information via iterative optimization. Besides the edge map used in~\citep{nazeri2019edgeconnect}, \cite{Yamashita2022Boundary} incorporated the depth image to provide the boundaries between different objects. Their method first completed the masked edge and depth images separately and then recovered the missing regions via an RGB image inpainting network taking as input the concatenation of masked images, inpainted edges, and depth images. To contain richer structural information, \cite{wu2022deep} choose the local binary pattern (LBP)~\citep{ojala1996a,ojala2002multi}, which describes the distribution information of edges, speckles, and other local features~\citep{zhang2010local}. In~\citep{wu2022deep}, the first network infers the LBP information of the holes, and the second network with spatial attention conducts the actual image inpainting. \cite{dong2022incremental} utilized a transformer to complete the holistic structure in a grayscale space and proposed a masking positional encoding for large irregular masks.

In addition, semantic segmentation maps are also used as the proxy of structure~\citep{song2018spg,qiu2021semantic,zhou2021image}. \cite{song2018spg} introduced the semantic segmentation information into the image inpainting process to improve the recovered boundary between different class regions. They first predict the segmentation map of missing regions via a U-Net and then recover the missing contents with the guidance of the above inpainted semantic map using the second generator network. \cite{song2018spg} utilized the pre-classification algorithm~\citep{graphseg} to extract a semantic structure map. After the completion of the semantic map, they employed a spatial-channel attention module to generate the texture information. \cite{zhou2021image} first predicted the complete segmentation map via a segmentation reconstructor, and then recovered fine-grained texture details with an image generator based on a relation network. The relation network is an extension of SPADE~\citep{spade} to better modulate features via spatially-adaptive normalization with the relation graph.

\begin{figure}[t]
    \centering
    \includegraphics[width=1.\linewidth]{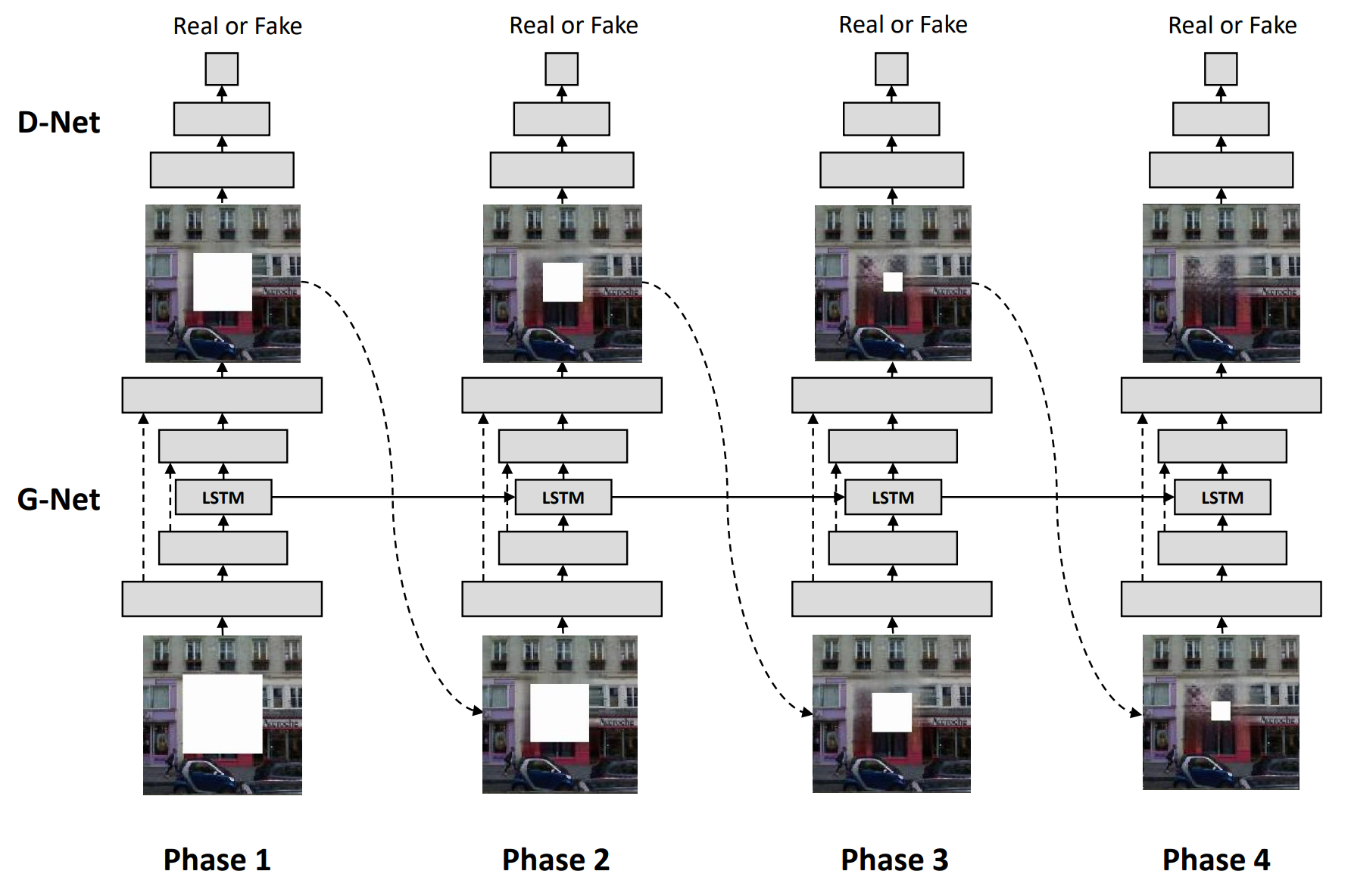}
    \caption{Progressive image inpainting. The image comes from~\citep{zhang2018semantic}.}
    \label{fig:progressive inpainting}
\end{figure}

\subsubsection{Progressive frameworks}
Following the basic idea of traditional inpainting methods, some works have been proposed to exploit progressive inpainting with deep models. As shown in Fig.~\ref{fig:progressive inpainting}, the progressive methods iteratively fill in the holes from the boundary to the center of the holes, and the missing area gradually becomes smaller until it disappears. \cite{zhang2018semantic} formulated image inpainting as a sequential problem, where the missing regions are filled in four inpainting phases. They designed an LSTM (long short-term memory)~\citep{sepp1997long}-based framework to string these four inpainting phases together. However, this method cannot handle irregular holes common in real-world applications. \cite{guo2019progressive} devised a residual architecture to progressively update irregular masks and introduced a full-resolution network to facilitate feature integration and texture reconstruction. Inspired by structure-guided inpainting methods~\citep{nazeri2019edgeconnect,xiong2019foreground}, \cite{li2019progressive} proposed a progressive reconstruction with a visual structure network to incorporate structure information into the visual features step by step, which can generate a more structured image. Progressive inpainting methods have the potential to fill in large holes, however, it is still difficult due to the lack of constraints on the hole center. To handle this drawback, \cite{li2020recurrent} designed a recurrence feature reasoning network with consistent attention and weighted feature fusion. This network recurrently infers and gathers the hole boundaries of the feature map so as to progressively strengthen the constraints for estimating internal contents. \cite{zeng2020high} proposed an iterative inpainting method with confidence feedback for high-resolution images. SRInpaintor~\citep{li2022SRInpaintor} combined super-resolution and the transformer in a progressive pipeline. It reasons about the global structure in low resolution, and progressively refines the texture details in high resolution. 

To this end, we organize the important and prevalent technical aspects for the network design, as shown in Table~\ref{tab:technique_point}.

\begin{table*}
  \caption{The summary of important techniques for deep learning-based image inpainting.}
  \label{tab:technique_point}
  \centering
  \footnotesize
  \setlength{\tabcolsep}{1pt} 
  \begin{tabular}{c|c|c}
    \toprule
    Aspects & Blocks  & Core idea \\
    \midrule
    \multirow{4}{*}{mask-aware convolution}  
    & Shepard interpolation~\citep{ren2015shepard} & translation variant interpolation \\
    & partial convolution~\citep{liu2018image} & convolution on valid regions \\
    & gated convolution~\citep{yu2019free} & adaptive gating \\
    & priority-guided partial convolution~\citep{wang2021parallel} & structure and texture priority \\
    \midrule
    \multirow{5}{*}{Attention}  
    & contextual attention~\citep{yu2018generative} & background patches with high \\
    & &  similarity to the coarse prediction \\
    & coherent semantic attention~\citep{liu2019coherent} & correlation between patches within the hole \\
    & multi-scale attention module~\citep{wang2019multi} & attention with two patch sizes \\
    & multi-scale attention uint~\citep{qin2021multi} & attention with four different dilation rates \\
    \midrule
    \multirow{5}{*}{Normalization}
        & region normalization~\citep{yu2020region} & spatial and region-wise \\
        & probabilistic context normalization~\citep{wang2020vcnet} & transfer mean and variance \\ 
        & regional composite normalization~\citep{Wang2021dynamic} & batch, instance, and layer normalization \\
    & point-wise normalization~\citep{zhu2021image} & mask-ware batch normalization \\
    & frequency region attentive normalization~\citep{zhu2021image} & align low- and high-frequency features \\
    \midrule
    \multirow{5}{*}{Discriminator}  
    & global discriminator~\citep{pathak2016context} & entire image \\
    & local discriminator~\citep{iizuka2017globally} & corrupted region \\
    & patch-based discriminator (PatchDis)~\citep{yu2019free} & eense local patches \\
    & conditional multi-scale discriminator~\citep{li2020deepgin} & PatchDis with two different scales \\
    & soft mask-guided PatchDis~\citep{zeng2022aggregated} & central parts of the missing regions  \\
  \bottomrule
\end{tabular}
\end{table*}

\subsection{Stochastic Image Inpainting}
Image inpainting is an underdetermined inverse problem. Therefore, multiple plausible solutions exist. We use the term stochastic image inpainting to refer to methods capable of producing multiple solutions with a random sampling process.

\textbf{VAE-based methods.} A variational autoencoder (VAE)~\citep{Kingma2014auto} is a generative model that combines an encoder and a decoder. The encoder learns an appropriated latent space and the decoder transforms sampled latent representations back into new data. \cite{zheng2019pluralistic} proposed a two-branch completion network, where the reconstructive branch models the prior distribution of missing parts and reconstructs the original complete image from this distribution. The generative branch infers the latent conditional prior distribution for the missing areas. This framework is optimized by balancing the variance of the conditional distribution and the reconstruction of the original training data. \cite{zheng2021Pluralistic} extended this work by estimating the distributions in a separate training stage and introducing the patch-level short-long term attention module. For stochastic fashion image inpainting, \cite{han2019finet} decomposed the inpainting process as the shape and appearance generation. The network design for these two generation tasks mainly adopts the VAE architecture. Based on a pre-trained VAE on facial images, \cite{tu2019facial} first searched for the possible set of solutions in the coding vector space for the corrupted image, and then recovers possible face images with the decoder of the VAE. \cite{zhao2020diverse} proposed an instance-guided conditional image-to-image translation network to learn conditional completion distribution. Specifically, they first encode the instance and masked images into two probability feature spaces, and then design a cross-semantic attention layer to fuse two feature maps. A decoder is finally used to generate the inpainted image. However, \cite{han2019finet} and \cite{zhao2020diverse} often suffer from distorted structures and blurry textures due to the joint optimization of structure and appearance. \cite{peng2021generating} designed a two-stage pipeline, where the first stage produces multiple coarse results with different structures based on a hierarchical vector quantized variational auto-encoder, and the second stage synthesizes the texture under the guidance of the discrete structural features. 

\textbf{GAN-based methods.} GAN~\citep{goodfellow2014generative} learns the data distribution via an adversarial process. A generator is applied to transform sampled Gaussian random noise into image space and a discriminator is used to differentiate the real sample and fake sample. Based on the premise that the degree of freedom increases from the hole boundary to the hole center, \cite{liu2021pd} introduced a spatially probabilistic diversity normalization to modulate the pixel generation with diversity maps. Considering that minimizing the classical reconstruction loss hampers the diversity of results, they also proposed a perceptual diversity loss that maximizes the distance of two generated images in the feature space. By combining the image-conditional and unconditional generative architectures, \cite{zhao2021comodgan} proposed a co-modulated GAN for large-scale image inpainting. Technically, they encode the incomplete input image into a conditional latent vector, which is then concatenated with the original style vector of StyleGAN2~\citep{Karras2020analyze}. To enhance the diversity and control of image inpainting, \cite{zeng2021feature} applied the patch matching from the training samples on the basis of coarse inpainted results. In particular, they designed the nearest neighbor-based pixel-wise global matching (from a single image) and compositional matching (from multiple images). Inspired by CoModGAN~\citep{zhao2021comodgan}, \cite{zheng2022image} proposed a cascaded modulation GAN, which combines the global modulation and the spatially-adaptive modulation in each scale of the decoder, and replaces the common convolution with fast Fourier convolution~\citep{ffc} in the encoder. To directly complete the high-resolution image, \cite{li2022mat} proposed a mask-aware transformer module with a dynamic mask updating as~\citep{liu2018image}. This module conducts non-local interactions only using partially valid tokens in a shifted-window manner~\cite{liu2021swin}. Following~\citep{chen2018on,Karras2019a}, they developed a style manipulation module for stochastic generations.

\textbf{Flow-based methods.} Normalizing Flows~\citep{Tabak2010density,Dinh2015nice,Rezende2015variational} are a generative method that constructs a complex probability distribution by assembling a sequence of invertible mappings. Inspired by Glow~\citep{Kingma2018glow} and its conditional extension~\citep{lugmayr2020srflow}, \cite{wang2022diverse} proposed a conditional normalizing flow network to learn the probability distribution of structure priors. Then, another generator is applied to produce the final complete image with rich texture.

\textbf{MLM-based methods.} To produce a stochastic structure in the missing region, \cite{yu2021diverse} and \cite{wan2021high} adopted a sequence prediction pipeline based on a masked language model (MLM). \cite{yu2021diverse} proposed a bidirectional and auto-regressive transformer as the low-resolution stochastic-structure generator, which predicts masked token (missing regions) via a top-$\mathcal{K}$ sampling strategy during inference. Then, a texture generator was applied to generate multiple inpainted results. Similarly, \cite{wan2021high} proposed a Transformer-CNN framework. They first apply a transformer training with MLM objective to produce a low-resolution image with pluralistic structures and some coarse textures, and then utilize an encoder-decoder network to enhance the local texture details of the high-resolution complete image.

\begin{figure*}[htp!]
\centering
\includegraphics[width=1.\linewidth]{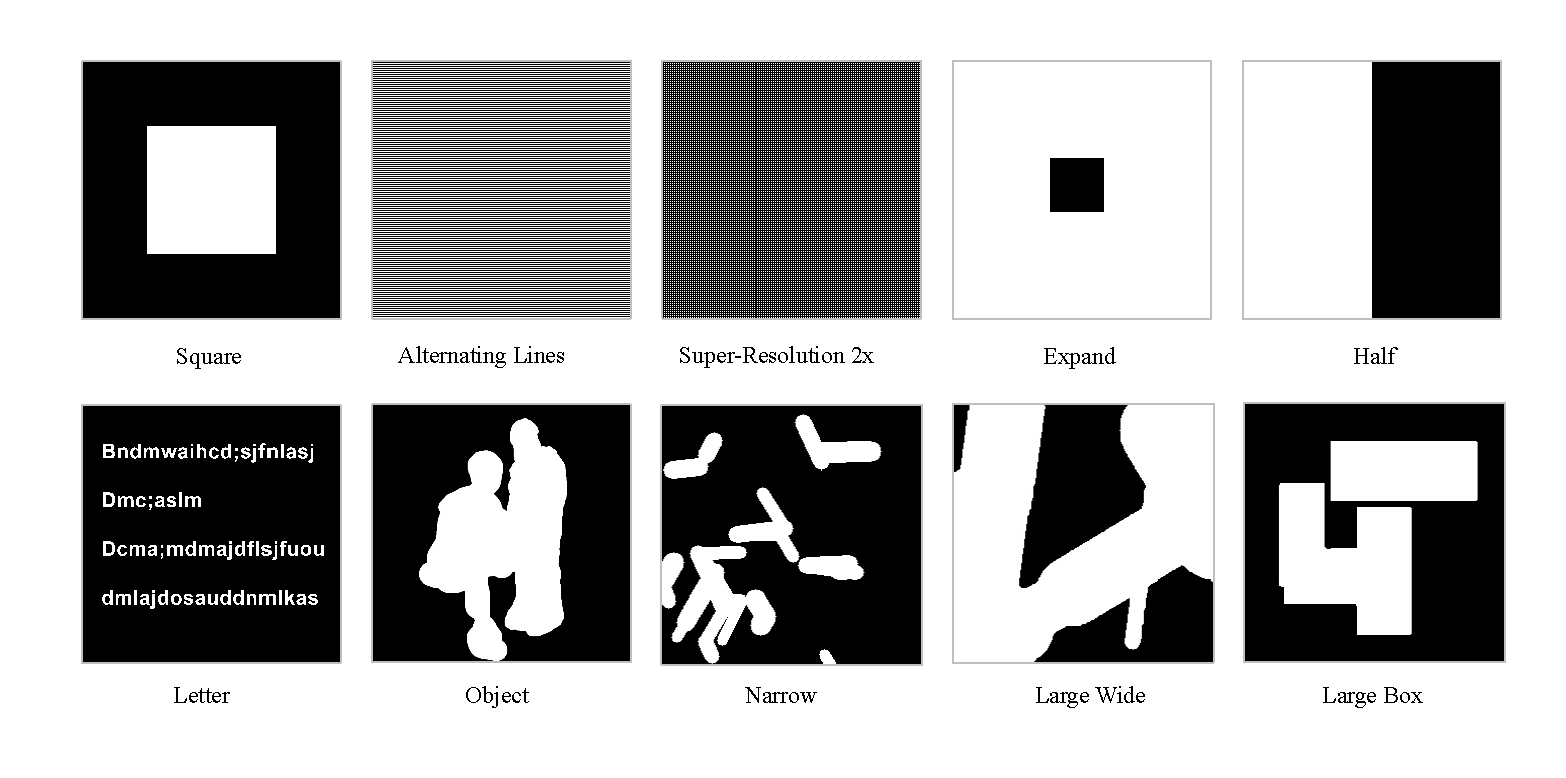}
\caption{Representative examples of masks.}
\label{fig:mask}
\end{figure*}

\textbf{Diffusion model-based methods.} Diffusion models (DM) are emerging generative models for image synthesis. Here, we only review diffusion model-based inpainting methods, and we refer readers to the surveys~\citep{yang2023diffusion,Croitoru2023diffusion} about a comprehensive introduction to diffusion models. Generally, diffusion-based inpainting models employ a U-Net architecture. The training objectives are usually  based on $\mathcal{L}_{DM}=\mathbb{E}_{x,\epsilon \in \mathcal{N}(0,1),t}[\|\epsilon-\epsilon_\theta(x_t,t)\|_2^2]$, where $t = 1 \dots T$, $x_t$ is a noised version of $x$, and $\epsilon_\theta(\cdot,t)$ is a neural network. In the literature, existing works mainly focused on the sampling strategy design and the computational cost reduction.  

(1) Sampling strategy design. \\
Based on an unconditionally pre-trained denoising diffusion probabilistic model (DDPM)~\citep{DDPM}, \cite{lugmayr2022RePaint} modified the standard denoising strategy by sampling the masked regions from the diffusion model and sampling the unmasked areas from the given image. To preserve the background and improve the consistency, \cite{Xie2023smart} added an extra mask prediction to the diffusion model. In the inference stage, the predicted mask is used to guide the sampling process.

(2) Computational cost reduction. \\
Instead of applying the diffusion process in pixel space, \cite{Esser2021ImageBART} utilized a multinomial diffusion process~\citep{Hoogeboom2021argmax,Austin2021structured} on a discrete latent space and autoregressively factorized models for the reverse process. These designs enable ImageBART to generate high-resolution images, \eg, $300 \times 1800$. Similarly, \cite{Rombach2022high} proposed a latent diffusion model (LDM) to reduce the training cost of DMs while boosting visual quality, which can be applied to the image inpainting task at a high resolution of $1024^2$ pixels. To overcome the limitation of massive iterations in the diffusion model, \cite{li2022sdm} proposed a spatial diffusion model (SDM) with decoupled probabilistic modeling, where the mean term refers to the inpainted result and the variance term measures the uncertainty. Instead of starting with random Gaussian noise in the reverse conditional diffusion, \cite{Chung2022come} remarkably reduced the number of sampling steps with a better initialization by starting from forward-diffused data.

\subsection{Text-guided Image Inpainting}
Text-guided image inpainting takes an incomplete image and text description as input and generates text-aligned inpainting results. The main challenge lies in how to fuse the text and image semantic features, and how to focus on effective information in the text. \cite{zhang2020text} proposed a dual attention mechanism to obtain the semantic feature of the masked region by finding unmatched words compared to the image and applied DAMSM loss~\citep{xu2018attngan} to measure the similarity of text and image. \cite{lin2020mmfl} introduced an image-adaptive word demand module that removes redundant information and aggregates text features in the coarse stage. They also proposed a text-guided attention loss that pays more attention to the reconstruction of the region affected by the text. \cite{zhang2020text-guided} encoded text and image to sequential data and exploited the transformer architecture to let cross-modal features interact. To ensure that the inpainted image matches the text, they took the masked text and inpainted image as input to restore the text prompt. \cite{wu2021adversarial} incorporated word-level and sentence-level textual features into a two-stage generator by introducing a dual-attention module. To eliminate the effection of the background, the mask reconstruction module was devised to recover the corrupted object mask. \cite{xie2022learning} applied multi-head self-attention as text-image interactive encoder. They created a semantic relation graph to compute non-Euclidean semantic relations between text and image, and used graph convolution to aggregate node features. \cite{li2023migt} followed a coarse-to-fine image inpainting framework. They first employed a visual-aware textual filtering mechanism to adaptively concentrate on required words and then inserted filtered text features into the coarse network. Unlike~\citep{zhang2020text-guided}, they directly reconstructed text descriptions from inpainted images to guarantee multi-modal semantic alignment. To better preserve the non-defective regions during the text guidance, \cite{ni2023nuwa} proposed a defect-free VQGAN to control receptive spreading and a sequence-to-sequence module to enable visual-language learning from multiple different perspectives, including text descriptions, low-level pixels, and high-level tokens. Recent methods are based on diffusion models.\cite{shukla2023scene} focused on how to generate a high-quality text prompt to guide a text-to-image model-based inpainting network by analyzing inter-object relationships. They first constructed a scene graph based on object detector outputs and expanded it via a graph convolution network to obtain the features of the corrupted node. Finally, the generated text prompt and masked image were fed to the diffusion model to obtain the inpainted result. \cite{wang2023imagen} found that object masks would force the inpainted images to rely more on text descriptions instead of the random mask. Then, they proposed Imagen Editor fine-tuned from Imagen~\citep{saharia2022photorealistic} with a new convolutional layer and designed an object masking strategy for better training. To facilitate the systematic evaluation of text-guided image inpainting, they established a benchmark called EditBench.

\subsection{Inpainting Mask}
In the development of image inpainting techniques, various artificial masks have been introduced. These masks can be roughly divided into two categories: regular masks and irregular masks. Fig.~\ref{fig:mask} summarizes these masks, where white pixels indicate missing regions.

\textbf{Regular masks.}
A \emph{square hole} that blocks the center area or random location are generally easier to construct. \cite{lugmayr2022RePaint} introduced more regular masks, including \emph{Super-Resolution $2\times$} (reserving pixels with a stride of 2), \emph{Alternating lines} (removing every second row), \emph{Expand} (leaving a small center crop of the input image), and \emph{Half} (masking the half of the input image).

\textbf{Irregular masks.}
\emph{Letter masks} (\citep{bertalmio2000image,Bian2022Scence}) and \emph{object-shaped masks} (\citep{criminisi2004region,yi2020contextual}) are particularly designed for specific tasks, for example, caption removal and object removal. \cite{liu2018image} introduced free-form masks, where the former collected random streaks and arbitrary holes from the results of the occlusion/dis-occlusion mask estimation method. The irregular masks shared by~\citep{liu2018image} are very common in the existing inpainting methods. \cite{lama} further split free-form masks into \emph{narrow masks}, \emph{large wide masks}, and \emph{large box masks}, where two types of large masks are generated via an aggressive mask method sampling polygonal chains with a high random width and rectangles of random aspect ratios, respectively.

\subsection{Loss Functions}
\label{subsec:loss funcions}
For image inpainting, the loss functions affect features of different sizes. At the lowest level, a pixel reconstruction loss aims to recover the exact pixel values. We further discuss the total-variational (TV) loss~\citep{Rudin1992nonlinear}, feature consistency loss, the perceptual loss~\citep{johnson2016perceptual}, style loss~\citep{Gatys2016image}, and adversarial loss~\citep{goodfellow2014generative}.

As input, an inpainting network accepts an input image $\mathbf{I}_{in}$ and a binary mask $\mathbf{M}$ describing the missing regions (where 0 means the valid pixel and 1 means the missing pixel). The output of the network is a complete image $\mathbf{I}_{out}$. The loss functions are formulated as follows.

\textbf{Pixel-wise reconstruction loss.} In the literature, the pixel-wise reconstruction loss often has two types: $\ell_1$ loss (Eq.~\eqref{equ:loss_pr0}) and weighted $\ell_1$ loss (Eq.~\eqref{equ:loss_pr}). The key point is how the valid and unknown regions differ in the loss function. The detailed formulations are as follows,
\begin{equation}
\mathcal{L}_{wpr} = ||(\mathbf{I}_{gt} - \mathbf{I}_{out}) ||_1.
\label{equ:loss_pr0}
\end{equation}
where $\mathbf{I}_{gt}$ is the ground-truth complete image.
\begin{equation}
\begin{split}
\mathcal{L}_{valid} & = \frac{1}{sum(\mathbbm{1} - \mathbf{M})}||(\mathbf{I}_{gt} - \mathbf{I}_{out}) \odot (\mathbbm{1} - \mathbf{M})||_1,  \\
\mathcal{L}_{hole} & = \frac{1}{sum(\mathbf{M})}||(\mathbf{I}_{gt} - \mathbf{I}_{out}) \odot \mathbf{M}||_1,
\end{split}
\end{equation}
where $\odot$ is the element-wise product operation, and $sum(\mathbf{M})$ is the number of non-zero elements in $\mathbf{M}$. Then the weighted $\ell_1$ loss is formulated as
\begin{equation}
    \mathcal{L}_{pr} = \mathcal{L}_{valid} + \alpha \cdot \mathcal{L}_{hole},
\label{equ:loss_pr}
\end{equation}
where $\alpha$ is the balancing factor. It is well known that the $\ell_1$ loss can capture the low-frequency components, whereas it struggles to restore the high-frequency components~\citep{isola2017image,srgan2017}.

\textbf{Total-variation loss.} Total-variation loss can be applied to ameliorate the potential checkerboard artifacts introduced by the perceptual loss. The formulation is:
\begin{equation}
\begin{split}
      \mathcal{L}_{tv} = ||\mathbf{I}_{mer}(i,j+1) - \mathbf{I}_{mer}(i,j)||_1 \\ 
      + ||\mathbf{I}_{mer}(i+1,j) - \mathbf{I}_{mer}(i,j)||_1.  
\end{split}
\end{equation}
where $\mathbf{I}_{mer} = \mathbf{I}_{out} \odot \mathbf{M} + \mathbf{I}_{gt} \odot (\mathbbm{1} - \mathbf{M})$ is the merged (completed) image.

\textbf{Feature consistency loss.} This loss constrains extracted feature maps of the prediction with guidance from ground truth images:
\begin{equation}
\begin{split}
      \mathcal{L}_{fc} = \sum_{y \in \Omega}||\Phi_m(\mathbf{I}_{in})_y - \Phi_n(\mathbf{I}_{gt})_y||_2^2.  
\end{split}
\end{equation}
where $\Omega$ is the missing regions, $\Phi_m(\cdot)$ is the feature map of the selected layer in the inpainting network, and $\Phi_n(\cdot)$ is the feature map of the corresponding layer in the inpainting network or pre-trained VGG models. $\Phi_m(\cdot)$ and $\Phi_n(\cdot)$ must have the same shape.

\begin{figure*}[ht]
\centering
\includegraphics[width=1.\linewidth]{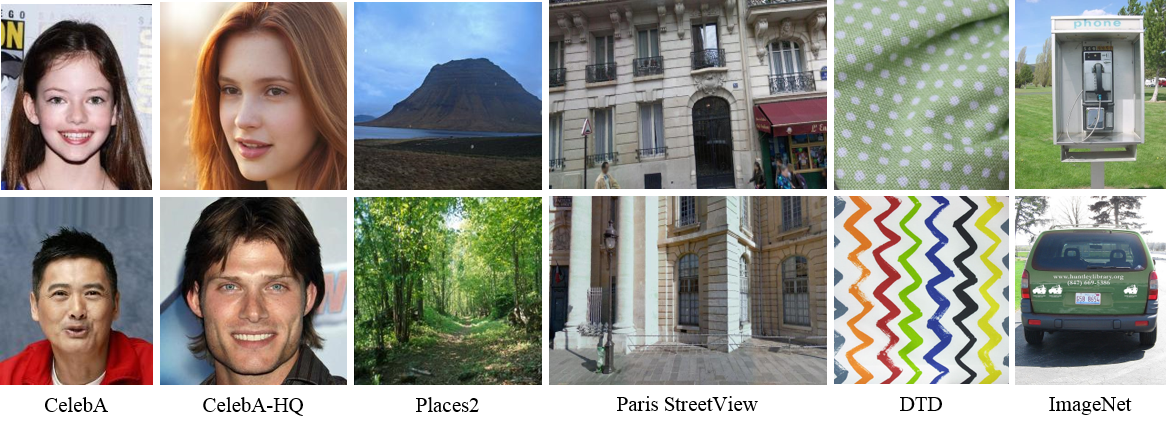}
\caption{Some examples of image inpainting datasets.}
\label{fig:inpainting_dataset}
\end{figure*}

\textbf{Perceptual loss.} The perceptual loss is first proposed in style transfer and super-resolution tasks. This loss measures the semantic/content difference between inpainted and ground-truth images, and thus encourages the inpainting generator to restore the semantics of missing regions. The perceptual loss is computed in high-level feature representations and is formulated as:
\begin{equation}
    \mathcal{L}_{per} = \sum_{i}||\Psi_i(\mathbf{I}_{out}) - \Psi_i(\mathbf{I}_{gt})||_1 + ||\Psi_i(\mathbf{I}_{mer}) - \Psi_i(\mathbf{I}_{gt})||_1,
\label{equ:loss_perceptual}
\end{equation}
where $\Psi_i(*)$ is the feature map of $i$-th layer in the VGG-16/19 network~\citep{simonyan2014very} pre-trained on ImageNet~\citep{deng2009imagenet}. Instead of using the common VGG network, \cite{lama} suggested using a base network with a fast-growing receptive field for large-mask inpainting and utilized the pre-trained segmentation network (ResNet50 with dilated convolutions~\citep{zhou2018semantic}) to compute the so-called high receptive field perceptual loss. Note that, some works used 2-norm in Eq.~\eqref{equ:loss_perceptual} to compute perceptual loss.

\textbf{Style loss.} Similar to the perceptual loss, the style loss also depends on higher-level features extracted from a pre-trained network. This loss is applied to penalize the style difference between inpainted and ground-truth images, \eg, texture details and common patterns. Mathematically, the style loss measures the similarities of Gram matrices of image features, instead of the feature reconstruction in the perceptual loss. The detailed formulation can be written,
\begin{equation}
    \mathcal{L}_{sty} = \sum_{i}||\Phi_i(\mathbf{I}_{out}) - \Phi_i(\mathbf{I}_{gt})||_1 + ||\Phi_i(\mathbf{I}_{mer}) - \Phi_i(\mathbf{I}_{gt})||_1,
\label{equ:loss_style}
\end{equation}
where $\Phi_i(\cdot) = \Psi_i(\cdot)\Psi_i(\cdot)^{T}$ is the Gram matrix~\citep{Gatys2016image}.

Besides using Gram matrices to model the style information, the mean and standard deviation of image features are commonly used in style transfer~\citep{huang2017arbitrary,deng2020arbitrary}. The formulation is written as, 
\begin{equation}
\begin{split}
    \mathcal{L}_{sty\_mean} = \sum_{i}||\mu(\Psi_i(\mathbf{I}_{out})) - \mu(\Psi_i(\mathbf{I}_{gt}))||_2 \\ 
    + ||\mu(\Psi_i(\mathbf{I}_{mer})) - \mu(\Psi_i(\mathbf{I}_{gt}))||_2, \\
    \mathcal{L}_{sty\_std} = \sum_{i}||\sigma(\Psi_i(\mathbf{I}_{out})) - \sigma(\Psi_i(\mathbf{I}_{gt}))||_2 \\ 
    + ||\sigma(\Psi_i(\mathbf{I}_{mer})) - \sigma(\Psi_i(\mathbf{I}_{gt}))||_2, \\
    \mathcal{L}_{sty\_meanstd} = \mathcal{L}_{sty\_mean} + \mathcal{L}_{sty\_std},
\end{split}
\label{equ:loss_style_meanstd}
\end{equation}
where $\mu(*)$, $\sigma(*)$ are the mean and standard deviation, computed over spatial dimensions independently for each sample and each channel.

\textbf{Adversarial loss.} GANs~\citep{goodfellow2014generative} are widely used in many image generation tasks. They employ an adversarial loss to force the output distribution to be close to the ``real'' distribution. The adversarial loss can counteract blurry results and enhance the visual realism of the output image. Therefore, it is often applied in GAN-based inpainting networks. To compute the adversarial loss, a discriminator network ($D$) is necessary, which interacts with the generator network ($G$). The hinge version~\citep{GeoGAN} of the adversarial loss can be formulated as:
\begin{equation}
\begin{split}
        \mathcal{L}_D = \mathbb{E}_{\mathbf{\mathbf{I}} \sim p_{data}(\mathbf{\mathbf{I}})} \Bigl[max(0, 1 - D(\mathbf{I}_{gt}))\Bigr] \\
        + \mathbb{E}_{\mathbf{\mathbf{I}_{mer}} \sim p_{\mathbf{I}_{mer}}(\mathbf{\mathbf{I}_{mer}})} \Bigl[max(0, 1 + D(\mathbf{I}_{mer}))\Bigr],
\end{split}
\label{equ:lsgan_dis}
\end{equation}
where $D(\mathbf{I}_{gt})$ and $D(\mathbf{I}_{mer})$ are the logits output from discriminator $D$. The objective function for generator $G$ can be denoted as:
\begin{equation}
        \mathcal{L}_G = -\mathbb{E}_{\mathbf{\mathbf{I}_{mer}} \sim p_{\mathbf{I}_{mer}}(\mathbf{\mathbf{I}_{mer}})} \Bigl[D(\mathbf{I}_{mer})\Bigr].
\label{equ:lsgan_gen}
\end{equation}
Except for the above hinge version, other types of adversarial losses are also adopted: GAN~\citep{goodfellow2014generative}, WGAN~\citep{WGAN2017}, LSGAN~\citep{mao2017least}, etc.

\subsection{Datasets}
In the literature, there are six prevalent and public datasets for evaluating image inpainting. These datasets cover various types of images, including faces (CelebA and CelebA-HQ), real-world encountered scenes (Places2), street scenes (Paris), texture (DTD), and objects (ImageNet). Several examples are shown in Fig.~\ref{fig:inpainting_dataset}. The details of the datasets are described below.

\begin{itemize}
    \item CelebA dataset~\citep{liu2015deep}: A large-scale face attribute dataset that contains 10,177 identities, each of which has about 20 images. In total, CelebA has 202,599 face images, each with 40 attribute annotations.
    
    \item CelebA-HQ dataset~\citep{karras2018progressive}: The high-quality version of CelebA~\citep{liu2015deep} with JPEG artifacts removal, super-resolution operation, and cropping, etc. This dataset consists of 30,000 face images.
    
    \item Places2 dataset~\citep{zhou2017places}: A large-scale scene recognition dataset. Places365-Standard has 365 scene categories. The training set has 1,803,460 images and the validation set contains 18,250 images.
    
    \item Paris StreetView dataset~\citep{doersch2012what}: This dataset consists of street-level imagery. It contains 14,900 images for training and 100 images for testing.
    
    \item DTD dataset~\citep{dtd2014}: A describable texture dataset consisting of 5,640 images. According to human perception, these images are divided into 47 categories with 120 images per category.
    
    \item ImageNet dataset~\citep{deng2009imagenet}: A large-scale benchmark for object category classification. There are about 1.2 million training images and 50 thousand validation images.
    
\end{itemize}

\setlength{\tabcolsep}{4pt}
\begin{table*}[htp!]
\begin{center}
\caption{Quantitative comparison of several representative image inpainting methods on CelebA-HQ and Places2. $\ddagger$ Higher is better. $\dagger$ Lower is better. From M1 to M6, the mask ratios are 1\%-10\%, 10\%-20\%, 20\%-30\%, 30\%-40\%, 40\%-50\%, and 50\%-60\%, respectively. Because of the heavy inference time, we do not show the results of RePaint for M1, M2, M4, and M6.}
\label{Tab:comparison_image}
\footnotesize
\setlength{\tabcolsep}{6pt} \centering
  \begin{tabular}{c|c||c|c|c|c|c|c||c|c|c|c|c|c}
    \hline
    & Dataset  & \multicolumn{6}{c||}{CelebA-HQ} & \multicolumn{6}{c}{Places2}  \\ \hline
    & Mask  & M1              & M2                 & M3              & M4                & M5              & M6                & M1              & M2                 & M3              & M4                & M5              & M6               \\ \hline \hline
    \multirow{8}{*}{\rotatebox{90}{$\ell_1(\%)$ $\dagger$}}
    & RFR & 1.59           & 2.47              & 3.58           & 4.90             & 6.44           & 9.47             & 0.83           & 2.20             & 3.93           & 5.83             & 7.96          & 11.37           \\ 
    & MADF   & 0.47           & 1.30              & 2.40           & 3.72             & 5.26           & 8.43             & 0.80           & 2.18             & 3.96           & 5.91            & 8.10          & 11.68           \\ 
    & DSI & 0.60  & 1.65     & 3.08           & 4.80             & 6.83           & 11.11             & 0.88           & 2.42             & 4.48           & 6.75            & 9.32          & 13.82           \\
    & CR-Fill & 0.79  & 2.15     & 3.95           & 6.01             & 8.33           & 13.18             & 0.78           & 2.17             & 4.02           & 6.11            & 8.46          & 12.43           \\
    & CoModGAN & 0.48  & 1.38     & 2.66           & 4.28             & 6.20           & 10.53             & 0.72           & 2.05             & 3.83           & 5.89            & 8.27          & 12.58           \\
    & LGNet & 0.46  & 1.28     & 2.38           & 3.72             & 5.27           & 8.38             & 0.68           & 1.89             & 3.51           & 5.33            & 7.41          & 10.86           \\
    & MAT  & 0.83           & 1.74     & 3.00  & 4.52    & 6.30  & 9.98    & 1.07  & 2.53    & 4.48  & 6.69    & 9.20 & 13.70  \\
    & RePaint  & -           & -     & 3.37  & -    & 7.47  & -    & -  & -    & 4.96  & -    & 10.01 & 15.27  \\ \hline \hline
    \multirow{8}{*}{\rotatebox{90}{PSNR $\ddagger$}}
        & RFR & 36.39           & 31.87              & 29.07           & 26.87             & 25.09           & 22.51             & 35.74           & 30.24             & 27.24           & 25.13             & 23.48          & 21.33           \\ 
    & MADF   & 39.68           & 33.77              & 30.42           & 27.95             & 25.99           & 23.07             & 36.17           & 30.37             & 27.17           & 25.00            & 23.31          & 21.10           \\ 
    & DSI & 37.68  & 31.74     & 28.39           & 25.88             & 23.91           & 20.87             & 35.40           & 29.47             & 26.15           & 23.91            & 22.19          & 19.75           \\
    & CR-Fill & 35.67  & 29.87     & 26.60           & 24.29             & 22.53           & 19.70             & 36.35           & 30.32             & 26.96           & 24.63            & 22.85          & 20.50           \\
    & CoModGAN & 39.56  & 33.15     & 29.41           & 26.62             & 24.49           & 21.16             & 37.00          & 30.82             & 27.35           & 24.92            & 23.05          & 20.43           \\
    & LGNet & 40.04  & 33.99     & 30.54           & 27.99             & 26.01           & 23.12             & 37.62           & 31.61             & 28.18           & 25.84            & 24.05          & 21.69           \\
    & MAT  & 38.44           & 32.62     & 29.21  & 26.70    & 24.72  & 21.78    & 35.66  & 29.76    & 26.41  & 24.09    & 22.30 & 19.81  \\
    & RePaint  & -           & -     & 28.38  & -    & 23.16  & -    & -  & -    & 26.04  & -    & 21.72 & 18.99  \\ \hline \hline
    \multirow{8}{*}{\rotatebox{90}{SSIM $\ddagger$}}
        & RFR & 0.991           & 0.976              & 0.957           & 0.932             & 0.902           & 0.834             & 0.983           & 0.952             & 0.911           & 0.862             & 0.805          & 0.699           \\ 
    & MADF   & 0.995           & 0.984              & 0.967           & 0.945             & 0.917           & 0.848             & 0.984           & 0.953             & 0.910           & 0.859            & 0.800          & 0.690           \\ 
    & DSI & 0.992  & 0.976     & 0.951           & 0.918             & 0.877           & 0.778             & 0.982           & 0.945             & 0.892           & 0.832            & 0.763          & 0.636           \\
    & CR-Fill & 0.988  & 0.965     & 0.931           & 0.890             & 0.842           & 0.729             & 0.985           & 0.954             & 0.909           & 0.855            & 0.794          & 0.675           \\
    & CoModGAN & 0.994  & 0.981     & 0.960           & 0.929             & 0.891           & 0.792             & 0.987           & 0.957             & 0.914           & 0.860            & 0.796          & 0.671           \\
    & LGNet & 0.995  & 0.985     & 0.968           & 0.945             & 0.917           & 0.849             & 0.988           & 0.963             & 0.925           & 0.878            & 0.823          & 0.714           \\
    & MAT  & 0.993           & 0.980     & 0.959  & 0.931    & 0.897  & 0.814    & 0.983  & 0.948    & 0.898  & 0.839    & 0.772 & 0.645  \\
    & RePaint  & -           & -     & 0.952  & -    & 0.867  & -    & -  & -    & 0.892  & -    & 0.750 & 0.606  \\ \hline \hline
    \multirow{8}{*}{\rotatebox{90}{MS-SSIM $\ddagger$}}
        & RFR & 0.992           & 0.976              & 0.956           & 0.933             & 0.900           & 0.830             & 0.986           & 0.960             & 0.924           & 0.880             & 0.828          & 0.731           \\ 
    & MADF   & 0.994           & 0.983              & 0.966           & 0.942             & 0.913           & 0.846             & 0.987           & 0.961             & 0.923           & 0.877            & 0.824          & 0.722           \\ 
    & DSI & 0.992  & 0.976     & 0.952           & 0.919             & 0.878           & 0.784             & 0.984           & 0.952             & 0.905           & 0.850            & 0.785          & 0.664           \\
    & CR-Fill & 0.987  & 0.963     & 0.928           & 0.887             & 0.839           & 0.732             & 0.987           & 0.960             & 0.920           & 0.872            & 0.814          & 0.704           \\
    & CoModGAN & 0.994  & 0.980     & 0.958           & 0.926             & 0.888           & 0.793             & 0.988           & 0.961             & 0.921           & 0.870            & 0.810          & 0.692           \\
    & LGNet & 0.995  & 0.984     & 0.968           & 0.945             & 0.917           & 0.851             & 0.990           & 0.968             & 0.935           & 0.894            & 0.844          & 0.744           \\
    & MAT  & 0.994           & 0.980     & 0.960  & 0.932    & 0.898  & 0.818    & 0.986  & 0.957    & 0.913  & 0.859    & 0.796 & 0.676  \\
    & RePaint  & -           & -     & 0.953  & -    & 0.870  & -    & -  & -    & 0.903  & -    & 0.771 & 0.633  \\ \hline \hline
    \multirow{8}{*}{\rotatebox{90}{FID $\dagger$}}
        & RFR & 0.86           & 1.68              & 2.67           & 3.77             & 5.21           & 7.60             & 2.62           & 5.99             & 9.47           & 12.90             & 16.62          & 22.13           \\ 
    & MADF   & 0.52           & 1.55              & 3.28           & 5.43             & 8.35           & 13.54             & 2.15           & 5.58             & 9.20           & 13.08            & 17.36          & 24.42           \\ 
    & DSI & 0.59  & 1.58     & 3.01           & 4.50             & 6.51           & 9.76             & 2.51           & 6.52             & 11.35           & 15.99            & 21.75          & 29.38           \\
    & CR-Fill & 1.06  & 2.86     & 5.26           & 7.79             & 11.23           & 19.52             & 2.37           & 6.24             & 10.54           & 15.17            & 20.36          & 26.43           \\
    & CoModGAN & 0.44  & 1.25     & 2.45           & 3.65             & 5.03           & 6.89             & 2.11           & 5.63             & 9.58           & 13.65            & 17.68          & 22.58           \\
    & LGNet & 0.39  & 1.06     & 2.08           & 3.16             & 4.61           & 7.07             & 1.97           & 5.25             & 8.90           & 13.02            & 17.60          & 25.99           \\
    & MAT  & 0.41           & 1.13     & 2.05  & 2.96    & 4.05  & 5.43    & 2.13  & 5.47    & 9.26  & 13.00    & 16.62 & 21.88  \\
    & RePaint  & -           & -     & 2.14  & -    & 4.24  & -    & -  & -    & 8.85  & -    & 15.90 & 21.58  \\ \hline \hline
    \multirow{8}{*}{\rotatebox{90}{LPIPS $\dagger$}}
        & RFR & 0.015           & 0.028              & 0.042           & 0.060             & 0.081           & 0.118             & 0.021           & 0.047             & 0.074           & 0.106             & 0.142          & 0.201           \\ 
    & MADF   & 0.009           & 0.025              & 0.048           & 0.077             & 0.109           & 0.168             & 0.014           & 0.038             & 0.068           & 0.102            & 0.141          & 0.209           \\ 
    & DSI & 0.010  & 0.026     & 0.048           & 0.074             & 0.104           & 0.160             & 0.018           & 0.047             & 0.085           & 0.125            & 0.169          & 0.242           \\
    & CR-Fill & 0.017  & 0.043     & 0.074           & 0.107             & 0.143           & 0.212             & 0.016           & 0.042             & 0.076           & 0.114            & 0.156          & 0.226           \\
    & CoModGAN & 0.008  & 0.022     & 0.041           & 0.065             & 0.092           & 0.143             & 0.016           & 0.044             & 0.080           & 0.121            & 0.164          & 0.236           \\
    & LGNet & 0.006  & 0.017     & 0.031           & 0.048             & 0.069           & 0.108             & 0.014           & 0.035             & 0.064           & 0.096            & 0.132          & 0.198           \\
    & MAT  & 0.007           & 0.019     & 0.035  & 0.054    & 0.077  & 0.120    & 0.014  & 0.040    & 0.073  & 0.111    & 0.152 & 0.224  \\
    & RePaint  & -           & -     & 0.038  & -    & 0.093  & -    & -  & -    & 0.077  & -    & 0.167 & 0.259  \\ \hline
  \end{tabular}
\end{center}
\end{table*}

\subsection{Evaluation Protocol}
The evaluation metrics can be classified into two categories: pixel-aware metrics and (human) perception-aware metrics. The former focus on the precision of reconstructed pixels, including $\ell_1$ error, $\ell_2$ error, and PSNR (peak signal-to-noise ratio), SSIM (the structural similarity index)~\citep{wang2004image}, and MS-SSIM (multi-scale SSIM)~\citep{wang2003multi}. The latter pay more attention to the visual perception quality, including FID (Fr{\'e}chet inception distance)~\citep{heusel2017gans}, LPIPS (learned perceptual image patch similarity)~\citep{zhang2018lpips}, P/U-IDS (paired/unpaired inception discriminative score)~\citep{zhao2021comodgan}, and user study results. The detailed descriptions are given in the following. 

\begin{itemize}
    \item $\ell_1$ error: The mean absolute differences between the complete image ($\mathbf{I}_c$) and the ground-truth image ($\mathbf{I}_g$).
    
    \item $\ell_2$ error: The mean squared differences between the complete image and the ground-truth image.
    
    \item PSNR: It is mainly used to measure the quality of reconstruction of the complete image. Its formulation is $\mathrm{PSNR} = 20 \cdot log_{10}(255) - 10 \cdot log_{10}(\mathrm{MSE})$, where $\mathrm{MSE}$ is the mean squared error between the complete image and the ground-truth image.
    
    \item SSIM: Instead of estimating absolute errors, SSIM measures the similarity in structural information by incorporating luminance masking and contrast masking. It is written as $\mathrm{SSIM}=\frac{(2\mu_{\mathbf{I}_c}\mu_{I_g}+c_1)(2\sigma_{\mathbf{I}_c\mathbf{I}_g}+c_2)}{(\mu_{\mathbf{I}_c}^2+\mu_{\mathbf{I}_g}^2+c_1)(\sigma_{\mathbf{I}_c}^2+\sigma_{\mathbf{I}_g}^2+c_2)},$ where $\mu$ and $\sigma$ refer to the average and the variance, respectively; and $c_1 = 0.01^2$ and $c_2 = 0.03^2$ are two variables to stabilize the division.

    \item MS-SSIM: \cite{Dosselmann2011a} illustrated that SSIM is very close to the windowed mean squared error and \cite{wang2003multi} highlighted the single-scale nature of SSIM as a drawback. As an alternative, MS-SSIM embraces more flexibility for image quality assessment. To compute the MS-SSIM, two input images are iteratively processed with low-pass filters and downsampled with a stride of 2 (in total, five scales). Then, the contrast comparison and structure comparison are computed at each scale and the luminance comparison is calculated at the last scale. These measurements are combined with appropriate weights~\citep{wang2003multi}.
    
    \item FID: The Fr{\'e}chet inception distance compares two sets of images. It computes a Gaussian with mean and covariance $(\mathbf{m},\mathbf{C})$ and a Gaussian $(\mathbf{m}_g,\mathbf{C}_g)$ from deep features of the set of completed images and the set of ground-truth images. Specifically, FID is defined as $\mathrm{FID} = \lVert \mathbf{m} - \mathbf{m}_g \rVert_2^2+ \mathrm{Tr}(\mathbf{C}+\mathbf{C}_g - 2(\mathbf{C}\mathbf{C}_g)^{\frac{1}{2}})$.
    
    \item LPIPS: The distance of multi-layer deep features of complete and ground-truth images. Let $\mathbf{F}_c,\mathbf{F}_g \in \mathbb{R}^{H_l \times W_l \times C_l}$ denote the channel-wise normalized features in the $l$-th layer, the LPIPS is given by $\mathrm{LPIPS} = \sum_{l}\frac{1}{H_l W_l}\sum_{h,w}\lVert \mathbf{W}_l \odot ({\mathbf{F}_c^l}_{hw}-{\mathbf{F}_g^l}_{hw})\rVert_{2}^{2}$, where $\mathbf{W}_l \in \mathbb{R}^{C_l}$ is the channel weight vector.
    
    \item P/U-IDS: The linear separability of complete and ground-truth images in a pre-trained feature space. Let $\phi(\cdot)$ denote the Inception v3 model mapping the image to the 2048D feature space, $f(\cdot)$ be the decision function of the SVM, the P-IDS is formulated as $\mathrm{P\text{-}IDS} = \mathrm{Pr}\{f(\phi(\mathbf{I}_c)) > f(\phi(\mathbf{I}_g)\}$. Due to the unpaired nature, U-IDS is obtained by directly calculating the misclassification rate.
    
    \item User Study: FID, LPIPS, and P/U-IDS cannot be able to comprehensively evaluate the visual quality of complete images, therefore, a user study is often conducted to complement the above metrics. User studies typically let a human chooses a preferred image among two (or multiple) images generated from two (or multiple) competitors. Based on the collected votes, the preference ratio is calculated for comparison. 
    
\end{itemize}

\begin{figure*}[t]
    \centering
    \includegraphics[width=1.\linewidth]{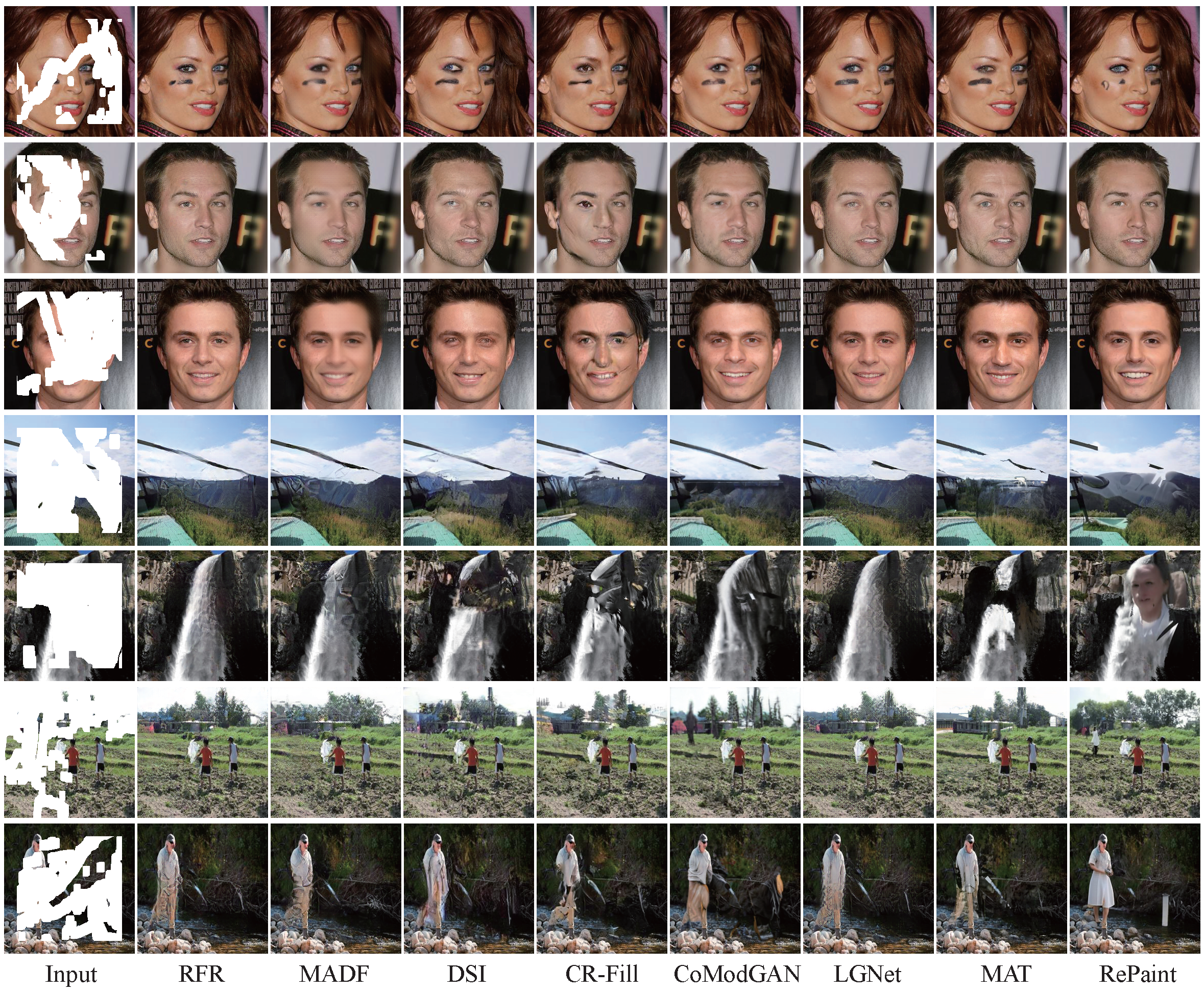}
    \caption{Qualitative comparison of representative image inpainting methods on CelebA-HQ (the first three rows) and Places2 (the last four rows).}
    \label{fig:compare_image_inpainting}
\end{figure*}

\begin{table}[t]
    \centering
    \caption{Model computational complexity statistics.}
    \setlength{\tabcolsep}{3pt}
    \begin{tabular}{c|c|c|c}
    \hline
    Model   & \#Parameter & GPU Memory & Infer. time \\ \hline
    RFR & 30.59 M & 1.23 G & 28.95 ms  \\ 
    MADF & 85.14 M & 2.42 G & 15.59 ms  \\   
    DSI  & 70.32 M & 6.54 G & 40.20 s  \\
    CR-Fill & 4.10 M & 0.96 G & 9.18 ms  \\
    CoModGAN & 79.80 M & 1.71 G & 42.24 ms  \\
    LGNet  & 115.00 M & 1.52 G & 13.59 ms  \\
    MAT & 59.78 M & 1.69 G & 78.35 ms  \\
    RePaint & 552.81 M & 4.14 G & 6 min 30 s \\ \hline
    \end{tabular}
    \label{tab:model_sta}
\end{table}

\setlength{\tabcolsep}{4pt}
\begin{table*}[t]
\begin{center}
\caption{Quantitative comparison of different loss functions on CelebA-HQ (``C'') and Paris StreetView (``P''). $\ddagger$ Higher is better. $\dagger$ Lower is better. ``16'' refers to Eq.~\eqref{equ:loss_pr} with $\alpha = 6$, and the remaining loss settings both include ``16'' (We omit it for simplicity). ``percept'' refers to Eq.~\eqref{equ:loss_perceptual} based on pretrained VGG16; ``resnetpl'' refers to Eq.~\eqref{equ:loss_perceptual} based on the pre-trained segmentation network ResNet50, which is proposed by~\citep{lama}.; ``style'' refers to Eq.~\eqref{equ:loss_style}; ``stylemeanstd'' refers to Eq.~\eqref{equ:loss_style_meanstd}; ``percept\_style'' refers to Eq.~\eqref{equ:loss_perceptual} plus Eq.~\eqref{equ:loss_style}; ``lsgan'' refers to Eq.~\eqref{equ:lsgan_dis} and~\eqref{equ:lsgan_gen}. Different percentage numbers in the first row refer to the hole ratios, where a large number implies large missing regions. Following the common setting, the test mask is from~\citep{liu2018image}.}
\label{Tab:comparison_loss}
\footnotesize
\setlength{\tabcolsep}{5pt} \centering
  \begin{tabular}{c|c||c|c||c|c||c|c||c|c||c|c||c|c}
    \hline
    & Mask  & \multicolumn{2}{c||}{1\%-10\%} & \multicolumn{2}{c||}{10\%-20\%} & \multicolumn{2}{c||}{20\%-30\%} & \multicolumn{2}{c||}{30\%-40\%} & \multicolumn{2}{c||}{40\%-50\%} & \multicolumn{2}{c}{50\%-60\%}  \\ \hline
    & Dataset  & C              & P                 & C              & P                & C              & P                & C              & P                & C              & P                & C              & P               \\ \hline \hline
    \multirow{7}{*}{\rotatebox{90}{$\ell_1(\%)$ $\dagger$}}
    & 16 & 0.46           & 0.57              & 1.26           & 1.53             & 2.34           & 2.81             & 3.63           & 4.25             & 5.14           & 5.93             & 8.37          & 9.07           \\ 
    & percept   & 0.45           & 0.57              & 1.24           & 1.53             & 2.30           & 2.81             & 3.58           & 4.26             & 5.08           & 5.95            & 8.33          & 9.11           \\ 
    & resnetpl & 0.45  & 0.59     & 1.25           & 1.58             & 2.32           & 2.88             & 3.60           & 4.35             & 5.10           & 6.06            & 8.35          & 9.21           \\
    & style & 0.48  & 0.59     & 1.33           & 1.60             & 2.47           & 2.97             & 3.85           & 4.53             & 5.45           & 6.37            & 8.86          & 9.78           \\
    & stylemeanstd & 0.46  & 0.59     & 1.27           & 1.60             & 2.39           & 2.96             & 3.75           & 4.51             & 5.35           & 6.34            & 8.80          & 9.75           \\
    & percept\_style & 0.47  & 0.60     & 1.30           & 1.63             & 2.43           & 3.00             & 3.79           & 4.58             & 5.40           & 6.42            & 8.83          & 9.84           \\
    & lsgan  & 0.47           & 0.59     & 1.30  & 1.61    & 2.42  & 2.97    & 3.76  & 4.52    & 5.33  & 6.34    & 8.67 & 9.72  \\ \hline \hline
    \multirow{7}{*}{\rotatebox{90}{PSNR $\ddagger$}}
        & 16 & 40.03           & 38.74              & 34.13           & 33.17             & 30.76           & 29.92             & 28.27           & 27.67             & 26.29           & 25.88             & 23.23          & 23.28           \\ 
    & percept   & 40.14           & 38.77              & 34.19           & 33.17             & 30.81           & 29.91             & 28.31           & 27.65             & 26.33           & 25.85            & 23.23          & 23.25           \\ 
    & resnetpl & 40.12  & 38.56     & 34.18           & 33.03             & 30.80           & 29.83             & 28.29           & 27.61             & 26.31           & 25.83            & 23.23          & 23.26           \\
    & style & 39.63  & 38.36     & 33.65           & 32.71             & 30.24           & 29.37             & 27.72           & 27.07             & 25.74           & 25.22            & 22.71          & 22.62           \\
    & stylemeanstd & 39.91  & 38.38     & 33.89           & 32.81             & 30.42           & 29.49             & 27.85           & 27.20             & 25.83           & 25.35            & 22.72          & 22.70           \\
    & percept\_style & 39.78  & 38.20     & 33.74           & 32.60             & 30.31           & 29.30             & 27.76           & 27.01             & 25.76           & 25.20            & 22.68          & 22.58           \\
    & lsgan  & 39.71           & 38.40     & 33.73  & 32.78    & 30.39  & 29.50    & 27.91  & 27.18    & 25.96  & 25.35    & 22.95 & 22.72  \\ \hline \hline
    \multirow{7}{*}{\rotatebox{90}{SSIM $\ddagger$}}
        & 16 & 0.995           & 0.991              & 0.985           & 0.973             & 0.969           & 0.946             & 0.948           & 0.911             & 0.921           & 0.867             & 0.847          & 0.767           \\ 
    & percept   & 0.995           & 0.991              & 0.985           & 0.973             & 0.970           & 0.946             & 0.949           & 0.911             & 0.921           & 0.866            & 0.847          & 0.765           \\ 
    & resnetpl & 0.995  & 0.991     & 0.985           & 0.972             & 0.970           & 0.945             & 0.948           & 0.909             & 0.921           & 0.865            & 0.848          & 0.764           \\
    & style & 0.995  & 0.991     & 0.983           & 0.971             & 0.966           & 0.940             & 0.943           & 0.902             & 0.913           & 0.853            & 0.834          & 0.746           \\
    & stylemeanstd & 0.995  & 0.991     & 0.984           & 0.971             & 0.968           & 0.941             & 0.944           & 0.903             & 0.914           & 0.854            & 0.835          & 0.747           \\
    & percept\_style & 0.995  & 0.990     & 0.984           & 0.970             & 0.967           & 0.940             & 0.943           & 0.901             & 0.913           & 0.852            & 0.834          & 0.745           \\
    & lsgan  & 0.995          & 0.991     & 0.984  & 0.971    & 0.967  & 0.941    & 0.944  & 0.903    & 0.915  & 0.854    & 0.839 & 0.746  \\ \hline \hline
    \multirow{7}{*}{\rotatebox{90}{FID $\dagger$}}
        & 16 & 0.56           & 4.74              & 1.57           & 13.74             & 3.31           & 26.55             & 5.38           & 40.79             & 8.37           & 57.49             & 15.18          & 86.51           \\ 
    & percept   & 0.53           & 4.64              & 1.51           & 13.41             & 3.20           & 26.13             & 5.22           & 40.35             & 8.18           & 57.22            & 14.63          & 88.10           \\ 
    & resnetpl & 0.52  & 4.62     & 1.47           & 13.23             & 3.11           & 25.60             & 5.13           & 39.08             & 7.99           & 54.75            & 13.81          & 83.93           \\
    & style & 0.42  & 3.91     & 1.13           & 10.67             & 2.25           & 19.65             & 3.38           & 28.87             & 5.00           & 39.09            & 7.90          & 57.00           \\
    & stylemeanstd & 0.44  & 4.21     & 1.21           & 11.42             & 2.38           & 20.54             & 3.65           & 29.68             & 5.36           & 39.59            & 8.55          & 56.38           \\
    & percept\_style & 0.40  & 3.98     & 1.13           & 10.91             & 2.26           & 19.76             & 3.42           & 29.12             & 5.07           & 39.28            & 7.87          & 57.07           \\
    & lsgan  & 0.54           & 4.26     & 1.57  & 11.72    & 3.34  & 21.54    & 5.57  & 31.21    & 8.85  & 41.80    & 16.03 & 60.01  \\ \hline \hline
    \multirow{7}{*}{\rotatebox{90}{LPIPS $\dagger$}}
        & 16 & 0.011           & 0.016              & 0.032           & 0.048             & 0.063           & 0.091             & 0.102           & 0.142             & 0.144           & 0.197             & 0.222          & 0.298           \\ 
    & percept   & 0.010           & 0.015              & 0.029           & 0.045             & 0.057           & 0.086             & 0.092           & 0.134             & 0.129           & 0.185            & 0.200          & 0.279           \\ 
    & resnetpl & 0.009  & 0.015     & 0.027           & 0.044             & 0.052          & 0.083             & 0.083           & 0.126             & 0.116           & 0.174            & 0.174          & 0.259           \\
    & style & 0.007  & 0.011     & 0.018           & 0.031             & 0.033           & 0.056             & 0.052           & 0.086             & 0.073           & 0.120            & 0.117          & 0.186           \\
    & stylemeanstd & 0.008  & 0.013     & 0.021           & 0.035             & 0.037           & 0.062             & 0.055           & 0.092             & 0.076           & 0.125            & 0.119          & 0.189           \\
    & percept\_style & 0.006  & 0.011     & 0.018           & 0.032             & 0.033           & 0.057             & 0.052           & 0.087             & 0.074           & 0.122            & 0.118          & 0.188           \\
    & lsgan  & 0.009           & 0.013     & 0.028  & 0.036    & 0.054  & 0.066    & 0.086  & 0.098    & 0.123  & 0.135    & 0.196 & 0.208  \\ \hline
  \end{tabular}
\end{center}
\end{table*}

\begin{figure*}[t]
\begin{center}
\includegraphics[width=1.\linewidth]{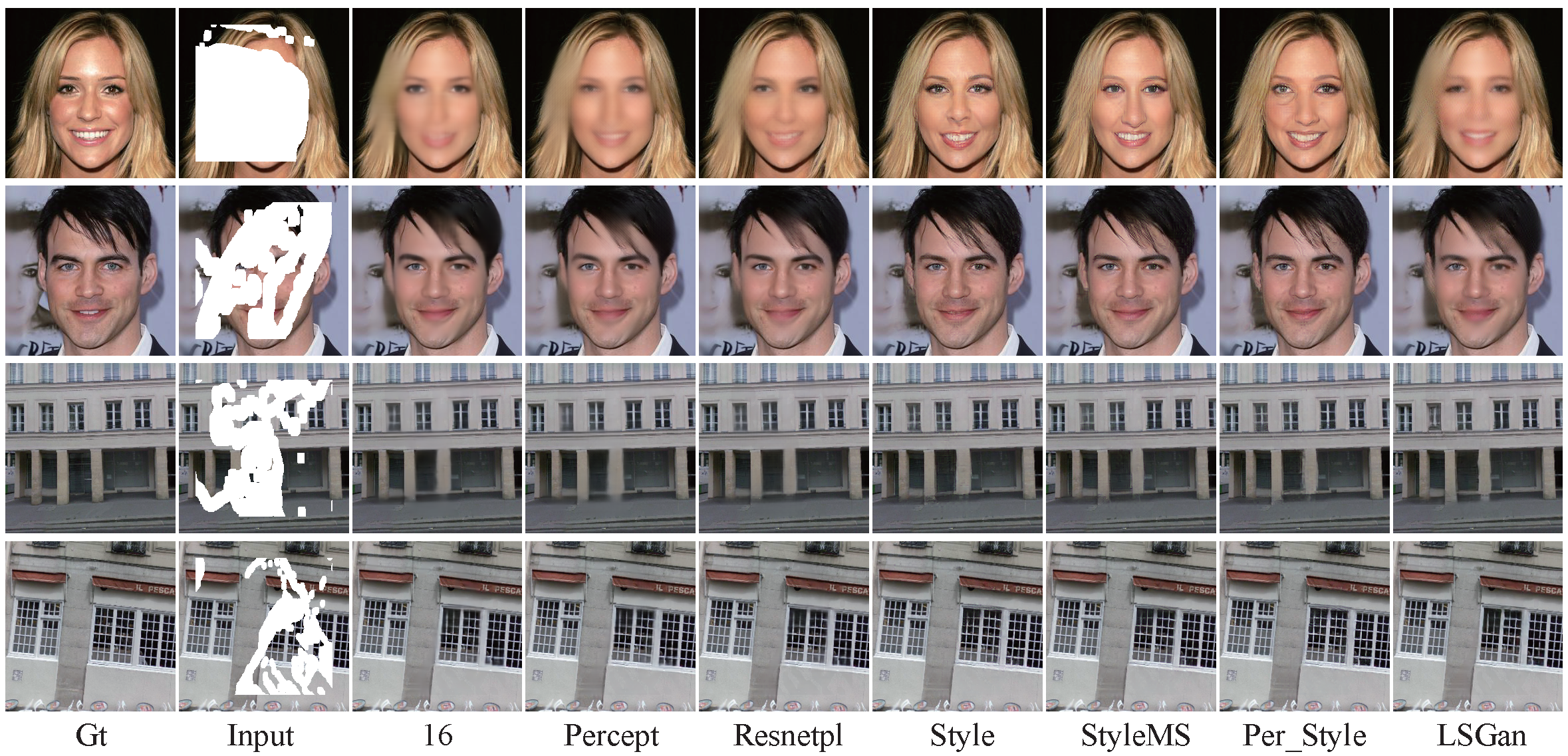}
\caption{Qualitative comparison of different loss functions on CelebA-HQ (the first two rows) and Paris StreetView (the last two rows). ``StyleMS'' refers to ``stylemeanstd''; ``Per\_Style'' refers to ``Percept\_style''.}
\label{fig:compare_loss}
\end{center}
\end{figure*}

\subsection{Performance Evaluation}

\subsubsection{Representative Image Inpainting Methods} We qualitatively and quantitatively compare some representative image inpainting methods: RFR~\citep{li2020recurrent}, MADF~\citep{zhu2021image}, DSI~\citep{peng2021generating}, CR-Fill~\citep{zeng2021cr}, CoModGAN~\citep{zhao2021comodgan}, LGNet~\citep{quan2022image}, MAT~\citep{li2022mat}, RePaint~\citep{lugmayr2022RePaint}. The test mask is from~\citep{liu2018image}. Specifically, RFR follows a progressive inpainting strategy, MADF adopts a mask-aware design, DSI generates stochastic structures with hierarchical vq-vae, CR-Fill designs an attention-free generator, CoModGAN embeds the known content of corrupted images into style vectors of styleGAN2, LGNet introduces local and global refinement networks with different receptive fields, MAT designs a mask-aware transformer architecture, and RePaint utilizes a pre-trained unconditional diffusion model.

Table~\ref{Tab:comparison_image} reports the quantitative results of these advanced image inpainting methods on CelebA-HQ and Places2 datasets. In this experiment, we use the irregular masks shared by~\citep{liu2018image} for the evaluation. From this table, we can find that MS-SSIM is very close to SSIM in the CelebA-HQ dataset; MS-SSIM is consistently higher than SSIM in the Places2 dataset and this phenomenon is more apparent for large masks. The reason may be that face images are relatively regular and uniform compared to the natural scene images in Places2, and thus the latter is more sensitive to structural similarity with different scales. Among these methods, MAT and RePaint have relatively superior FID, especially for large masks ($> 30\%$), while CoModGAN and LGNet perform better in PSNR. For DSI, the inpainting performance on CelebA-HQ is slightly better than that on Places2, and the possible reason is that the structure of face images is easier to model than diverse natural scene images. CR-Fill has limited inpainting performance. 

Fig.~\ref{fig:compare_image_inpainting} shows the visual results of some representative image inpainting methods on CelebA-HQ and Places2 datasets. MADF adopts a mask-aware design, which can predict reasonable structures (the second and third rows), but has limited ability for detail restoration. By contrast, MAT has better inpainting performance with mask-aware transformer blocks. Through introducing local and global refinement with different receptive fields, LGNet can perceive local details (the black stroke in the first row) and global structure (the second row). For large missing regions, RFR can recover the helicopter rotor blade (the fourth row) and waterfall (the fifth row) with progressive inpainting. With the help of the generative capability of the unconditional modulated model (StyleGAN2), CoModGAN demonstrates relatively good inpainting performance (the fourth and sixth rows). DSI can perceive the structure with hierarchical VQ-VAE (the third and fourth rows). Based on the powerful generation ability of the diffusion model, RePaint can correctly infer the missing background (the sixth row) and the human body (the seventh row). Interestingly, it may have an incorrect semantic prediction (a woman's head in the waterfall of the fifth row). Due to the implicit attention mechanism and simple network, CR-Fill achieves comparatively inferior inpainted results, which is also consistent with the quantitative comparisons as shown in Table~\ref{Tab:comparison_image}.

In addition, we evaluate the computational complexity of the representative inpainting methods in terms of the number of parameters, GPU memory of single image inference, and inference time on a GPU (the time of a forward pass through the networks. The statistical results are shown in Table~\ref{tab:model_sta}. CR-Fill implicitly learns the patch-borrowing behavior without an attention layer, its model is the smallest and thus needs less GPU memory and running time. Because RePaint is based on a diffusion model, it has the largest number of parameters and a very long inference time. The GPU memory and inference time of DSI are also very high. LGNet follows a coarse-to-fine framework with local and global refinement, therefore, the number of parameters is high. The running time of MAT and CoModGAN is relatively high because the former conducts many attention computations and the latter has multiple style modulations with progressive growing. RFR and MADF are in the middle.

\subsubsection{Loss Functions}
As summarized in Sec.~\ref{subsec:loss funcions}, many loss functions have been proposed for image inpainting. In this part, we evaluate the effect of each loss term. We train an inpainting network with different loss settings on the CelebA-HQ and Paris StreetView datasets. This network consists of two downsampling layers, 11 ResNet residual blocks with dilation, and two upsampling layers. The corresponding numerical results are reported in Table~\ref{Tab:comparison_loss}. In the case of masks at 1\%-10\%, the SSIM values of different loss settings are (almost) the same for CelebA-HQ and Paris StreetView datasets. The reason is that different loss settings only have a slight impact on the inpainting of very small missing regions. We can see that pixel-wise reconstruction loss (``16'') provides the baseline performance. After adding the perceptual loss (``percept''), FID and LPIPS are improved. Compared with ``percept'', ``resnetpl'' achieves slightly better results, especially for the large mask. The style loss can remarkably decrease the FID and LPIPS at the expense of PSNR and SSIM. In other words, there exists a trade-off between pixel-wise reconstruction loss and style loss, where the former focuses on low-level pixel recovery, and the latter emphasizes visual quality. A similar finding is reported and studied in~\citep{blau2018the}. In addition, combining the perceptual loss with style loss (``percept\_style'') has a very slight effect on the results compared to only style loss (``style''). The style loss based on Gram matrix (``style'') and style loss based on mean and standard deviation (``stylemeanstd'') have comparable results. Comparing with adversarial loss (``lsgan''), style loss (``style'') obtain significantly lower FID and LPIPS. 

Fig.~\ref{fig:compare_loss} illustrates the corresponding qualitative comparison. ``16'' fills the missing regions with smooth structures and textures. After introducing the content loss, this phenomenon is slightly improved, for example, the nose and mouth of the first row are better recovered in column ``Percept''. Compared with ``Percept'', the inpainted results of ``Resnetpl'' have slightly improved visual quality, which is attributed to the perceptual loss computation with higher receptive field~\citep{lama}. We can find that the results of ``Style'' are significantly superior to the previous three columns, especially for the restoration of texture details. This is consistent with the numerical results in Table~\ref{Tab:comparison_loss}. For three settings with style loss, \ie, ``Style'', ``StyleMS'', and ``Per\_Style'', ``Style'' and ``Per\_Style'' are on par, ``StyleMS'' is slightly worse. The performance of ``LSGan'' is in between ``Percept'' and ``Style''.

\subsection{Inpainting-based Applications}
Image inpainting can be used in many real-world applications, such as object removal, text editing, old photo restoration, image compression, text-guided image editing, etc.


\begin{figure}[t]
\begin{center}
\includegraphics[width=1.\linewidth]{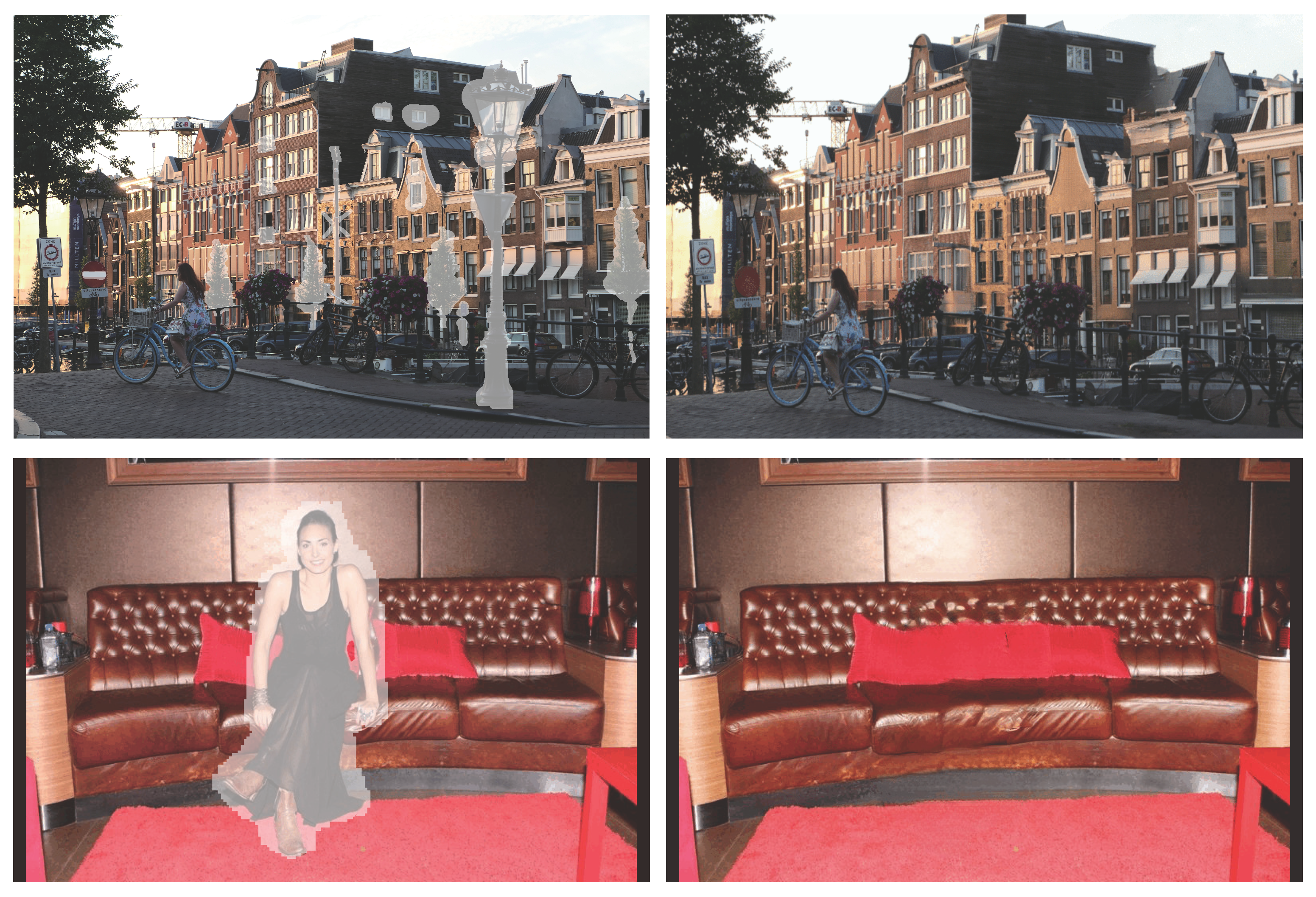}
\caption{Two representative examples of object removal.}
\label{fig:object_removal}
\end{center}
\end{figure}

\subsubsection{Object Removal} 
Almost all image editing tools include the function of object removal, which is directly accomplished with image inpainting. To illustrate the capability of several current inpainting methods on the object removal application, we apply the respective trained models to remove objects from selected real-world images with different scenes, and the corresponding results are shown in Fig.~\ref{fig:object_removal}. The first row is generated by CNN-based method~\citep{lama} and the second one is inpainted by a transformer-based method~\citep{zheng2022bridging}. These two methods can achieve visually realistic results, successfully removing the objects highlighted with shadow markers.

\begin{figure}[t]
    \centering
    \includegraphics[width=1.\linewidth]{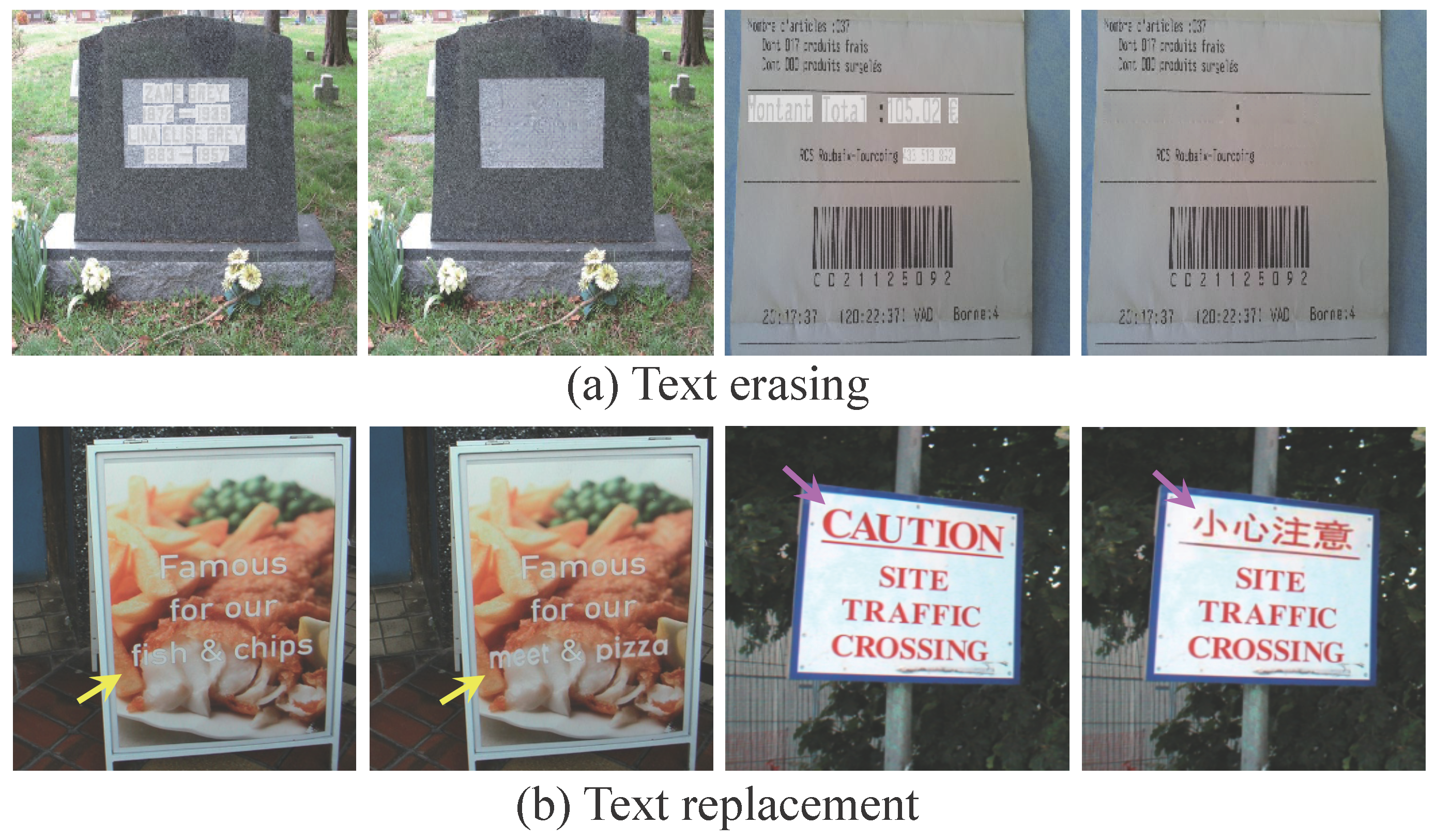}
    \caption{Representative samples of text editing.}
    \label{fig:text_editing}
\end{figure}

\subsubsection{Text Editing}
On social media sites, users often share their pictures and also want to hide their personal information for privacy. For real-time text translation applications in smartphones, the original content needs to be replaced with the translated version. These text editing-related tasks can be solved via inpainting techniques. Fig.~\ref{fig:text_editing}(a) shows the results of text removal with the method proposed by~\citep{quan2022image}, and Fig.~\ref{fig:text_editing}(b) illustrates two samples of text replacement from ~\citep{wu2019editing}. These results have a pleasing visual quality.

\subsubsection{Old Photo Restoration} 
\begin{figure}[t]
\begin{center}
\includegraphics[width=\linewidth]{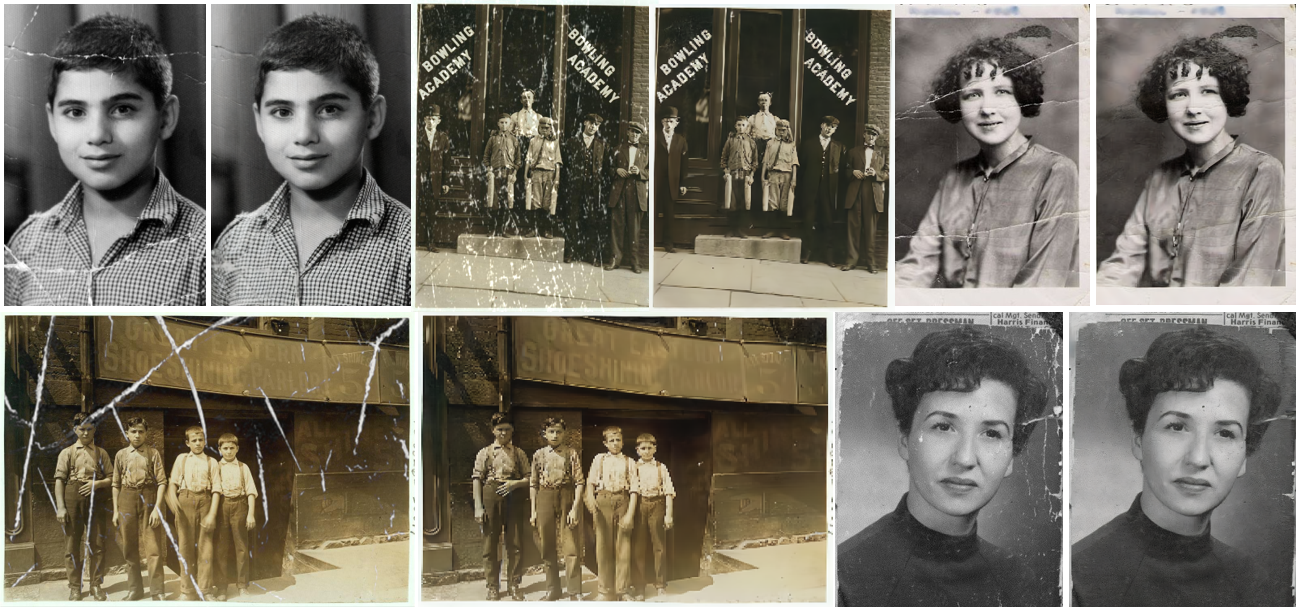}
\caption{Several representative examples of old photo restoration. }
\label{fig:oldphoto}
\end{center}
\end{figure}

Photos are helpful to record important moments. Unfortunately, some photos are damaged over time, resulting in various missing regions. Image inpainting can be used to recover these incomplete photos automatically. It is difficult to collect paired training data for this task, therefore, we synthesize old photos using the Pascal VOC dataset~\cite{voc} inspired by~\cite{wan2020bringing}. Specifically, we collect some paper and scratch texture images to simulate the realistic defects in the old photos. To blend the above texture images with the VOC images, we randomly choose a mode from three candidates (screen, lighten-only, and layer addition) with a random opacity. In addition, some operations, \eg, random flipping, random position, rescaling, cropping, etc, are also used for augmenting the diversity of texture images. To this end, the paired samples of the original VOC images and the corresponding blended results are used for training the inpainting network~\citep{quan2022image}. Fig.~\ref{fig:oldphoto} shows several examples of old photo restoration, where the inpainting method restores the original appearance of old photos.

\subsubsection{Image Compression} 
\begin{figure}[t]
    \begin{center}
    \includegraphics[width=1.\linewidth]{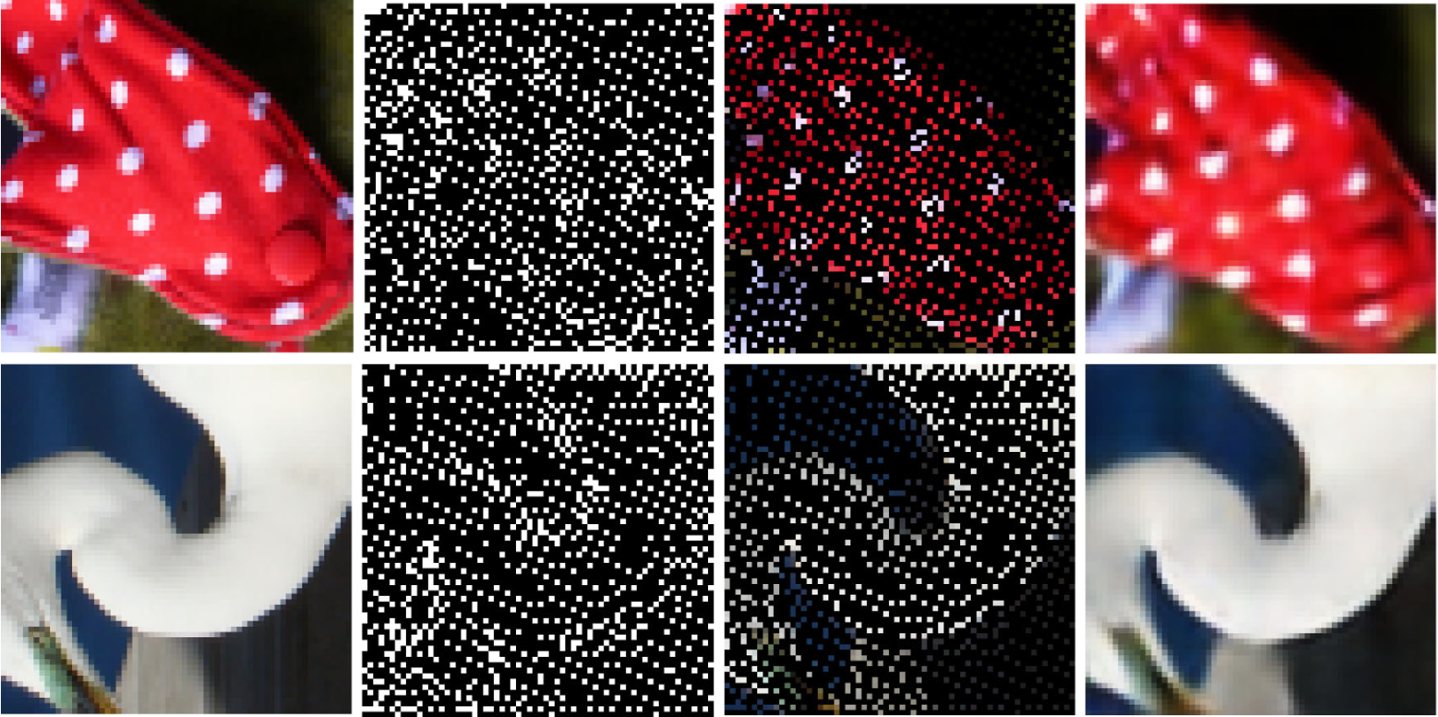}
    \caption{Two representative examples of image compression with inpainting. From left to right: input image, sampling mask, sampled image, and reconstructed image. Images come from~\citep{dai2020adaptive}. }
    \label{fig:image_compression}
    \end{center}
\end{figure}

Image compression is a fundamental image processing technique to reduce the cost of storage or transmission of digital images. This technique mainly consists of two stages: compression and reconstruction. The former reduces the data size to obtain a sparse image representation and the latter reconstructs the original image. Different from waveform-based methods, \cite{Carlsson1988Sketch} proposed a sketch-based method to obtain a sparse representation and reconstructed images via an interpolation process. \cite{Irena2008image} introduced partial differential equation (PDE)-based inpainting to image compression, where image coding and decoding both are based on edge-enhancing anisotropic diffusion. Recently, some researchers~\citep{Baluja2019Learning,dai2020adaptive,Schrader2023efficient} applied deep learning methods to generate the sampling mask and reconstruct the image with an inpainting network. Fig.~\ref{fig:image_compression} shows several examples of image compression with inpainting, where the reconstructed images have good quality based on adaptive sparse sampling with inpainting.

\subsubsection{Text-guided image editing}
\begin{figure}[t]
    \begin{center}
    \includegraphics[width=1.\linewidth]{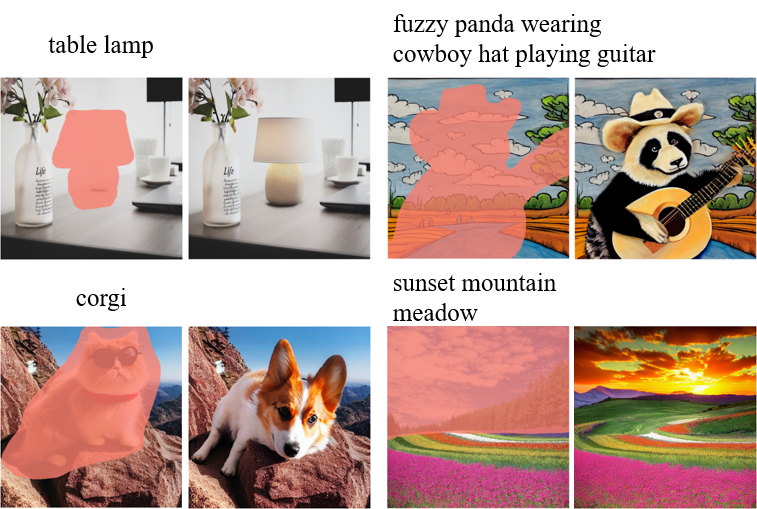}
    \caption{Selected examples of text-based image editing. Each group includes the input image with mask (red transparent) and text prompts and edited results. Images come from~\citep{Xie2023smart}.}
    \label{fig:text_guided_editing}
    \end{center}
\end{figure}

Image inpainting is a basic processing tool for image editing. Recent generative models based on probabilistic diffusion have the powerful capability of text-to-image generation, which provides the potential for text-guided image editing with diffusion model-based image inpainting approaches. For example, diffusion-based SmartBrush~\citep{Xie2023smart} edited images with the guidance of text and shape. Fig.~\ref{fig:text_guided_editing} illustrates several samples generated by SmartBrush. The first row adds new objects and the second row replaces original objects with new contents. We can see that the edited results have high visual realism and are consistent with text prompts. 

\section{Video Inpainting}
\subsection{Method}
Unlike images, videos have an additional temporal dimension which provides extra information about objects or camera movement. This information helps networks to obtain a better understanding of the context of the video. Therefore, the video inpainting task aims to ensure both spatial consistency and temporal coherence. Existing deep learning-based video inpainting methods can be roughly divided into four categories: 3D CNN-based approaches, shift-based approaches, flow-guided approaches, and attention-based approaches. We refer the readers to more conventional methods in~\citep{ilan2015a}.

\subsubsection{3D CNN-based Approaches} 
To deal with the temporal dimension, researchers proposed 3D CNN-based approaches, which often combine temporal restoration and image inpainting. \cite{wang2019video} proposed a two-stage pipeline to jointly infer temporal structure and spatial texture details. The first sub-network processes the low-resolution videos with a 3D CNN, and the second sub-network completes the original-resolution video frames with an extended 2D inpainting network~\citep{iizuka2017globally}. Inspired by the gated convolution in image inpainting~\citep{yu2019free}, \cite{chang2019free} proposed a 3D gated convolution and a temporal SN-PatchGAN for free-form video inpainting. They also integrated the perceptual loss~\citep{johnson2016perceptual} and style loss~\citep{Gatys2016image} into the training objective. \cite{hu2020proposal} proposed a two-stage video inpainting network, where they obtain a coarse inpainting result with a 3D CNN and then fuse inpainting proposals generated by matching valid pixels and pixels in coarse inpainting results.

\subsubsection{Shift-based Approaches}
Considering the high computational cost of 3D convolution, \cite{lin2019tsm} proposed a generic temporal shift module (TSM) to capture temporal relationships with high efficiency. This technique is extended for video inpainting. \cite{chang2019learnable} developed a learnable gated TSM, which combines a TSM with learnable shifting kernels and gated convolution~\citep{yu2019free}. They also equipped the 2D convolution layers in SN-PatchGAN~\citep{yu2019free} with gated TSM. However, TSM often leads to blurry content due to misaligned features. To solve this, \cite{zou2021progressive} proposed a spatially-aligned TSM (TSAM), aligning features to the current frame after shifting features. The alignment process is based on estimated flow with a validity mask. \cite{ouyang2021internal} applied an internal learning strategy for video inpainting, which implicitly learns the information shift from valid regions to unknown parts in a single video sample. They also designed the gradient regularization term and the anti-ambiguity loss term for temporal consistency reconstruction and realistic detail generation. \cite{ke2021occlusion} presented an occlusion-aware video object inpainting method. Specifically, they completed the object shape with a transformer-based network, recovered the flow within the completed object region under the guidance of the object contour, and filled missing content with an occlusion-aware TSM after the flow-guided pixel propagation.  

\subsubsection{Flow-guided Approaches} 
Optical flow is a common tool to model the temporal information in videos, which is also applied to solve video inpainting. Based on the completed flow, the missing pixels in the current frame can be filled by propagating pixels from neighboring frames. 
\cite{kim2019deep} modeled the video inpainting task as a multi-to-single frame inpainting problem and proposed a 3D-2D encoder-decoder network VINet. This network includes several flow and mask sub-networks in a progressive manner. They also introduced the flow and warp loss to further enforce temporal consistency. \cite{chang2019vornet} proposed a three-stage video inpainting framework consisting of a warping network, an inpainting network, and a refinement network. In the warping network, bilinear interpolation is used to recover background flow without learning. Then the refinement network selected the best candidate from two frames completed by warping and inpainting network to generate the final output. \cite{zhang2019internal} applied internal learning to infer both frames and flow from input random noise and used flow generation loss to enhance temporal coherence. \cite{xu2019deep} proposed a flow-guided completion framework consisting of three steps. It first fills the incomplete optical flow with stacked CNN networks, then propagates pixels from known regions to holes with inpainted flow guidance, and finally completes unseen regions with an image inpainting network~\citep{yu2019free}. To reduce the over-smoothing in the boundary regions during flow completion, they leveraged hard flow example mining to encourage the network to produce sharp edges. To solve the same problem, \cite{gao2020flow} explicitly completed motion edges and used them to guide flow completion. In addition, they introduced a non-local flow connection to enable content propagation from distant frames. 

These previous methods cannot guarantee the consistency of flow, and even small errors in the flow may lead to geometric distortion in the video. Inspired by this, \cite{lao2021flow} transformed the background of a 3D scene to a 2D scene template and learned the mapping of the template to the mask in the image. Given that the complex motion of objects between consecutive frames will increase the difficulty to recover flow, \cite{zhang2022inertia} introduced an inertia prior in flow completion to align and aggregate flow features. To alleviate the spatial incoherence problem, they proposed an adaptive style fusion network to correct the distribution in the warped regions with the guidance of feature distribution in valid regions. \cite{kang2022error} offset the weaknesses of the error accumulation of a multi-stage pipeline in flow-based methods by introducing an error compensation strategy, which iteratively detects and corrects the inconsistency errors during the flow-guided pixel propagation.

The above hand-crafted flow-based methods restored videos with high computation and memory consumption because these processes cannot be accelerated by GPU. To speed up training and inference, \cite{li2022towards} proposed an end-to-end framework. They propagated features based on completed flow in low resolution and used deformable convolution to decrease the distortion caused by errors in flow. The temporal focal transformer blocks were stacked to aggregate local and non-local features. 

\subsubsection{Attention-based Approaches} The attention mechanism is often applied to model the contextual information and enlarge the spatial-temporal window. \cite{oh2019onion} recurrently calculated the attention scores between the target and reference frames, and progressively filled holes of the target frame from the boundary. \cite{lee2019copy} firstly aligned frames by an affine transformation, and then copied pixels based on the similarity between the target frame and aligned reference frames. \cite{woo2020align} proposed a coarse-to-fine framework for video inpainting. The first stage roughly recovers the target holes based on the computed homography between the target and reference frames, and the second stage refines the filled contents with non-local attention. They also introduced an optical flow estimator to enhance temporal consistency. Considering the motion of the foreground objects is diverse, the choice of reference frames becomes more important. While other methods take neighboring frames or frames in a specific distance as reference frames, \cite{li2020short} dynamically updated long-term reference frames after aggregating short-term aligned features. 

\begin{table*}[t]
\caption{Summary of video inpainting methods. Like image inpainting, we also split existing video inpainting approaches into three types according to the number of stages: 1) one-stage framework (\Circled[inner color=blue]{1}) usually designs a generator to recover the missing contents for each frame; 2) two-stage framework (\Circled[inner color=green]{2}) often consists of two networks for different purposes; and 3) multi-stage framework (\Circled[inner color=red]{m}) splits video inpainting into multiple steps. }
    \label{tab:video_inpainting_summary}
    \setlength{\tabcolsep}{4pt}
    \centering
    \begin{tabular}{*{10}{c}}
        \toprule
        \multirow{2}{*}{Category} & \multirow{2}{*}{Method} &
        \multirow{2}{*}{Stage} & 
        \multicolumn{7}{c}{Loss details} \\
        \cmidrule(lr){4-10}
        & & & L1 loss & GAN loss & Perceptual loss & Style loss & TV loss & Flow loss & Warp loss \\
        \hline
        \hline
        \multirow{3}{*}{\rotatebox{90}{3D CNN}} 
        & \cite{wang2019video} & \Circled[inner color=green]{2} & \checkmark & & &  & \\
        & \cite{chang2019free} & \Circled[inner color=blue]{1} & \checkmark & \checkmark &  \checkmark & \checkmark & & & \\
        & \cite{hu2020proposal} & \Circled[inner color=green]{2} & \checkmark & \checkmark & & & \\
        \hline
        \hline
        \multirow{4}{*}{\rotatebox{90}{Shift}} 
        & \cite{chang2019learnable} & \Circled[inner color=blue]{1} & \checkmark & \checkmark & \checkmark & \checkmark & & \\
        & \cite{zou2021progressive} & \Circled[inner color=blue]{1} & \checkmark & \checkmark &  \checkmark & \checkmark & & \\
        & \cite{ouyang2021internal} & \Circled[inner color=blue]{1} & \checkmark &  & &  &  \\
        & \cite{ke2021occlusion} & \Circled[inner color=red]{m} & \checkmark & \checkmark & &  & & \checkmark & \checkmark  \\
        \hline
        \hline
        \multirow{9}{*}{\rotatebox{90}{Flow}} 
        & \cite{kim2019deep} & \Circled[inner color=blue]{1} & \checkmark & & & & & \checkmark & \checkmark \\
        & \cite{chang2019vornet} & \Circled[inner color=red]{m} & \checkmark & \checkmark & \checkmark & & & & \\
        & \cite{zhang2019internal} & \Circled[inner color=blue]{1} & \checkmark & & \checkmark & & & \checkmark & \checkmark \\
        & \cite{xu2019deep} & \Circled[inner color=red]{m} & & & & & & \checkmark & \\
        & \cite{gao2020flow} & \Circled[inner color=red]{m} & \checkmark & & & & & \checkmark & \\
        & \cite{lao2021flow} & \Circled[inner color=green]{2} & \checkmark & & & & & & \checkmark \\
        & \cite{zhang2022inertia} & \Circled[inner color=red]{m} & \checkmark & \checkmark & & & & \checkmark & \checkmark \\
        & \cite{li2022towards} & \Circled[inner color=blue]{1} & \checkmark & \checkmark & & & & \checkmark & \\
        & \cite{kang2022error} & \Circled[inner color=red]{m} & \checkmark & \checkmark & & & & \checkmark &  \\
        \hline
        \hline
        \multirow{9}{*}{\rotatebox{90}{Attention}} 
        & \cite{oh2019onion} & \Circled[inner color=red]{m} & \checkmark & & \checkmark & \checkmark & \checkmark \\
        & \cite{lee2019copy} & \Circled[inner color=green]{2} & \checkmark &  & \checkmark & \checkmark &\checkmark  & \\
        & \cite{woo2020align} & \Circled[inner color=green]{2} & \checkmark & \checkmark & & & & \checkmark & \checkmark \\
        & \cite{li2020short} & \Circled[inner color=blue]{1} & \checkmark & & \checkmark & \checkmark & & & \\
        & \cite{zeng2020learning} & \Circled[inner color=blue]{1} & \checkmark & \checkmark & &  & & & \\
        & \cite{liu2021fuseformer} & \Circled[inner color=blue]{1} & \checkmark & \checkmark & &  & & & \\
        & \cite{chen2021deep} & \Circled[inner color=green]{2} & \checkmark &  & \checkmark & \checkmark  & & & \\
        & \cite{zhang2022flow} & \Circled[inner color=green]{2} & \checkmark & \checkmark & & & & \checkmark &  \checkmark \\
        & \cite{ren2022dlformer} & \Circled[inner color=green]{2} & \checkmark &  & & & \checkmark & & \\
        \bottomrule
    \end{tabular}
\end{table*}

Instead of a frame-by-frame inpainting strategy, \cite{zeng2020learning} adopted a ``multi-to-multi'' mechanism to fill in the holes in all input frames. Specifically, they proposed a spatial-temporal transformer network (STTN) to compute attention in both spatial and temporal dimensions. Based on STTN~\citep{zeng2020learning}, \cite{liu2021fuseformer} separated feature maps into overlapping patches, enabling more interactions between neighboring patches. In addition, they modified the common transformer block by inserting soft split and soft composition modules into the feed-forward network. \cite{chen2021deep} proposed an interactive video inpainting method to jointly perform object segmentation and video inpainting with user guidance. For network design, they introduce a spatial time attention block to update the target frames' features with the reference frames' features. \cite{zhang2022flow} designed a flow-guided transformer to combine the flow and the attention. They first utilized the completed flow to propagate pixels from neighboring frames, and then synthesized the remaining missing regions with a flow-guided spatial transformer and a temporal transformer. 

These attention-based methods still suffer from blurry content in high frequency due to mapping videos into a continuous feature space. By learning a specific codebook for each video and using subscripts of code to represent images, \cite{ren2022dlformer} transformed videos to a discrete latent space. Then a discrete latent transformer was applied to infer content in masked regions.

Table~\ref{tab:video_inpainting_summary} summarizes the technical details of existing video inpainting methods.

\begin{table*}[t]
\caption{Quantitative comparisons of representative video inpainting methods on YouTube-VOS and DAVIS dataset. $\ddagger$ Higher is better. $\dagger$ Lower is better. *: our results using the method described in STTN~\citep{zeng2020learning}, and numerical differences may be due to different optical flow models during evaluation.}
    \label{tab:video_inpainting_sota}
    \setlength{\tabcolsep}{4pt}
    \centering
    \begin{tabular}{c|cccc|cccc}
        \toprule
        \multirow{2}{*}{Methods} &
        \multicolumn{4}{c|}{YouTube-VOS} & \multicolumn{4}{c}{DAVIS} \\
        \cmidrule(lr){2-5} \cmidrule(lr){6-9}
         & PSNR$\ddagger$ & SSIM$\ddagger$ & VFID$\dagger$ & FWE($\times 10^{-2}$)$\dagger$ & PSNR$\ddagger$ & SSIM$\ddagger$ & VFID$\dagger$ & FWE($\times 10^{-2}$)$\dagger$ \\
        \hline
        \hline
        VINet~\citep{kim2019deep} & 29.20 & 0.9434 & 0.072 & 0.1490 / - & 28.96 & 0.9411 & 0.199 & 0.1785 / - \\
        DFVI~\citep{xu2019deep} & 29.16 & 0.9429 & 0.066 & 0.1509 / - & 28.81 & 0.9404 & 0.187 & 0.1880 / 0.1608* \\
        LGTSM~\citep{chang2019learnable} & 29.74 & 0.9504 & 0.070 & 0.1859 / - & 28.57 & 0.9409 & 0.170 & 0.2566 / 0.1640* \\
        CAP~\citep{lee2019copy} & 31.58 & 0.9607 & 0.071 & 0.1470 / - & 30.28 & 0.9521 & 0.182 & 0.1824 / 0.1533* \\
        FGVC~\citep{gao2020flow} & 29.68 & 0.9396 & 0.064 & - / 0.0858* &  30.24 & 0.9444 & 0.143 & - / 0.1530*\\
        STTN~\citep{zeng2020learning} & 32.34 & 0.9655 & 0.053 & 0.1451 / 0.0884* &  30.67 & 0.9560 & 0.149 & 0.1779 / 0.1449*\\
        FuseFormer~\citep{liu2021fuseformer} & 33.16 & 0.9673 & 0.051 & - / 0.0875* & 32.54 & 0.9700 & 0.138 & - / 0.1336* \\
        FGT~\citep{zhang2022flow} & 32.11 & 0.9598 & 0.054 & - / 0.0860* & 32.39 & 0.9633 & 0.1095 & - / 0.1517*  \\
        ISVI~\citep{zhang2022inertia} & 32.80 & 0.9611 & 0.048 & - / 0.0856* & 33.70 & 0.967 & 0.1028 & - / 0.1509*  \\
        E2FGVI~\citep{li2022towards} & 33.50 & 0.9692 & 0.046 & - / 0.0864* & 32.71 & 0.9700 & 0.096 & - / 0.1383* \\
        \bottomrule
    \end{tabular}
\end{table*}

\subsection{Loss Functions}
Video inpainting is very close to image inpainting. Therefore, many loss functions for training image inpainting networks are also applied to train video inpainting models, including reconstruction loss, GAN loss, perceptual loss, and style loss. To complete the corrupted flow, two losses are often used: 

\textbf{Flow loss.} Similar to the image reconstruction loss, the flow loss measures the difference between inpainted flow and its ground-truth version, which is defined as:
\begin{equation}
    \mathcal{L}_{flow} = ||O_{i,j}\odot(F_{i,j}-\hat{F}_{i,j})||_1,
\label{equ:loss_flow}
\end{equation}
where $\hat{F}_{i,j}$ is the inpainted optical flow from frame $i$ to frame $j$, $F_{i,j}$ is the ground-truth flow estimated by pre-trained flow estimation networks, \eg, FlowNet2~\citep{FlowNet2} and PWC-Net~\citep{sun2018pwc}, and $O_{i,j}$ denotes the occlusion map obtained by the forward-backward consistency check.

\textbf{Warp loss.} This loss encourages image-flow consistency: 
\begin{equation}
    \mathcal{L}_{warp} = ||I_{i}-I_{j}(\hat{F}_{i,j})||_1,
\label{equ:loss_warp}
\end{equation}
where $I_{j}(\hat{F}_{i,j})$ refers to the warped result of the frame $I_{j}$ using the generated flow $\hat{F}_{i,j}$ through backward warping.

\begin{figure*}[t]
\begin{center}
\includegraphics[width=1.\linewidth]{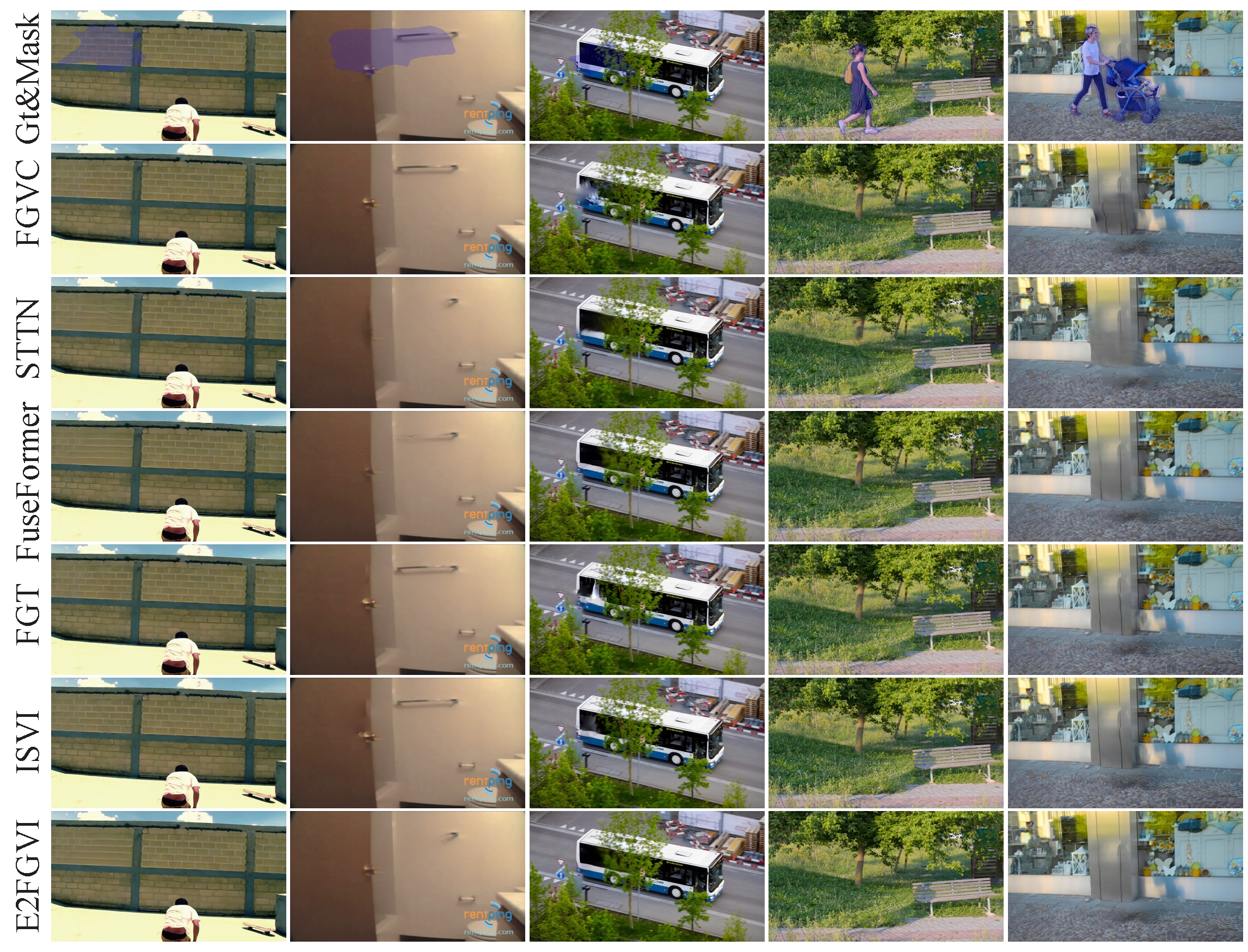}
\caption{Qualitative comparisons of representative video inpainting methods on YouTube-VOS and DAVIS dataset. The light blue mask highlights the corrupted regions. The first three columns are random masks and the remaining two columns are object masks. }
\label{fig:video_inpainting_sota}
\end{center}
\end{figure*}

\subsection{Datasets}
For video inpainting, three common video datasets, \ie, FaceForensics~\citep{roessler2018faceforensics}, DAVIS~\citep{perazzi2016a} and YouTube-VOS~\citep{xu2018youtube}, are used for training and evaluation. 

\begin{itemize}
   \item FaceForensics: A face forgery detection video dataset consisting of 1,004 videos. Among them, 854 videos are used for training and the rest are used for evaluation.
   
    \item DAVIS dataset: A densely annotated video segmentation dataset contains 150 videos with challenging motion-blur and appearance motions. For the data split, 60 videos are used for training and 90 videos for testing.

    \item YouTube-VOS dataset: A large-scale video object segmentation dataset containing 4,453 video clips and 94 object categories. The video clips have on average 150 frames and show various scenes. The original data split, \ie, 3,471/474/508, is adopted for experimental comparisons. 
\end{itemize}

\subsection{Evaluation Protocol}
Video contains many image frames, therefore, the two most widely-used metrics in image inpainting (\ie, PSNR and SSIM) are also used for video quality assessment. In addition, there are two other video-specific metrics (considering the temporal aspect), \ie, flow warping error (FWE)~\cite{lai2018learning} and video-based Fr{\'e}chet inception distance (VFID)~\cite{wang2018vid2vid}. The former evaluates the temporal stability of inpainted videos and the latter measures the perceptual realism in the video setting.

\begin{itemize}
    \item FWE: The flow warping error between two consecutive video frames is calculated as $\mathcal{E}(\mathbf{I}_t,\mathbf{I}_{t+1}) = \frac{1}{\sum_{n=1}^{N}\mathbf{M}_t^n}\sum_{n=1}^{N}\mathbf{M}_t^n||\mathbf{I}_t^{n}-\hat{\mathbf{I}}_{t+1}^{n}||^{2}_{2}$, where $\mathbf{M}_t$ is a binary mask indicating non-occluded areas and $\hat{\mathbf{I}}_{t+1}$ is the warped frame of $\mathbf{I}_{t+1}$. The non-occlusion mask can be estimated by using the method~\cite{ruder2016artistic}. Then, the warping error of a video is defined as the average error over the entire frames, and the formulation is $\mathcal{E} = \frac{1}{T-1}\sum_{t=1}^{T-1}\mathcal{E}(\mathbf{I}_t,\mathbf{I}_{t+1})$.
    
    \item VFID: A variant of FID for video evaluation. Instead of using a pre-trained image recognition network, the spatiotemporal feature map of each video is extracted via a pre-trained video recognition network, \eg, I3D~\cite{I3D}. Then, the VFID is calculated following the same procedure as the FID.
    
\end{itemize}

\subsection{Performance Evaluation}
In this section, we report the performance evaluation of representative video inpainting methods. 

Table~\ref{tab:video_inpainting_sota} shows the numerical results on YouTube-VOS and DAVIS datasets. We use the evaluated masks shared by~\citep{liu2021fuseformer}. Early video inpainting methods based on 3D convolution (\eg, VINet~\citep{kim2019deep}) and shift (\eg, LGTSM~\citep{chang2019learnable}) have relatively limited inpainting performance. After introducing optical flow and attention mechanisms, the quality of video inpainting is remarkably improved. DFVI~\citep{xu2019deep} generates the baseline result with flow guidance, and FGVC~\citep{gao2020flow} achieves better performance by completing flow with sharp edges and propagating information from distant frames. ISVI~\citep{zhang2022inertia} obtains more exact flow completion under the inertia prior, and thus enhances the inpainting quality. STTN~\citep{zeng2020learning} and FuseFormer~\citep{liu2021fuseformer} both design video inpainting frameworks through stacking multiple transformer blocks with multi-scale attention and dense patch-wise attention, respectively. FGT~\citep{zhang2022flow} and E2FGVI~\citep{li2022towards} combine the flow completion and transformer as a whole, and the end-to-end pipeline as adopted by E2FGVI is slightly better.

Fig.~\ref{fig:video_inpainting_sota} illustrates some inpainted results with different types of scenes and masks. From the first and second columns, we find that the flow-based pixel propagation methods, including FGVC, FGT, and ISVI, have a good ability to recover the texture details and objects with the guidance of neighboring frames. Through contextual correlation modeling, transformer-based video inpainting methods, such as STTN, FuseFormer, and E2FGVI, can complete the structure of objects, \eg, the window of a bus in the third column. Compared to STTN, FuseFormer introduces more dense attention computation (with overlapping), which can help the global structure recovery, \eg, the trunk in the fourth column and the post in the last column. In the fourth column, the coverage area is better filled with the realistic grass texture by the ISVI method, which is attributed to the more accurate flow completion compared to FGVC and FGT.

\begin{figure}[t]
\begin{center}
\includegraphics[width=\linewidth]{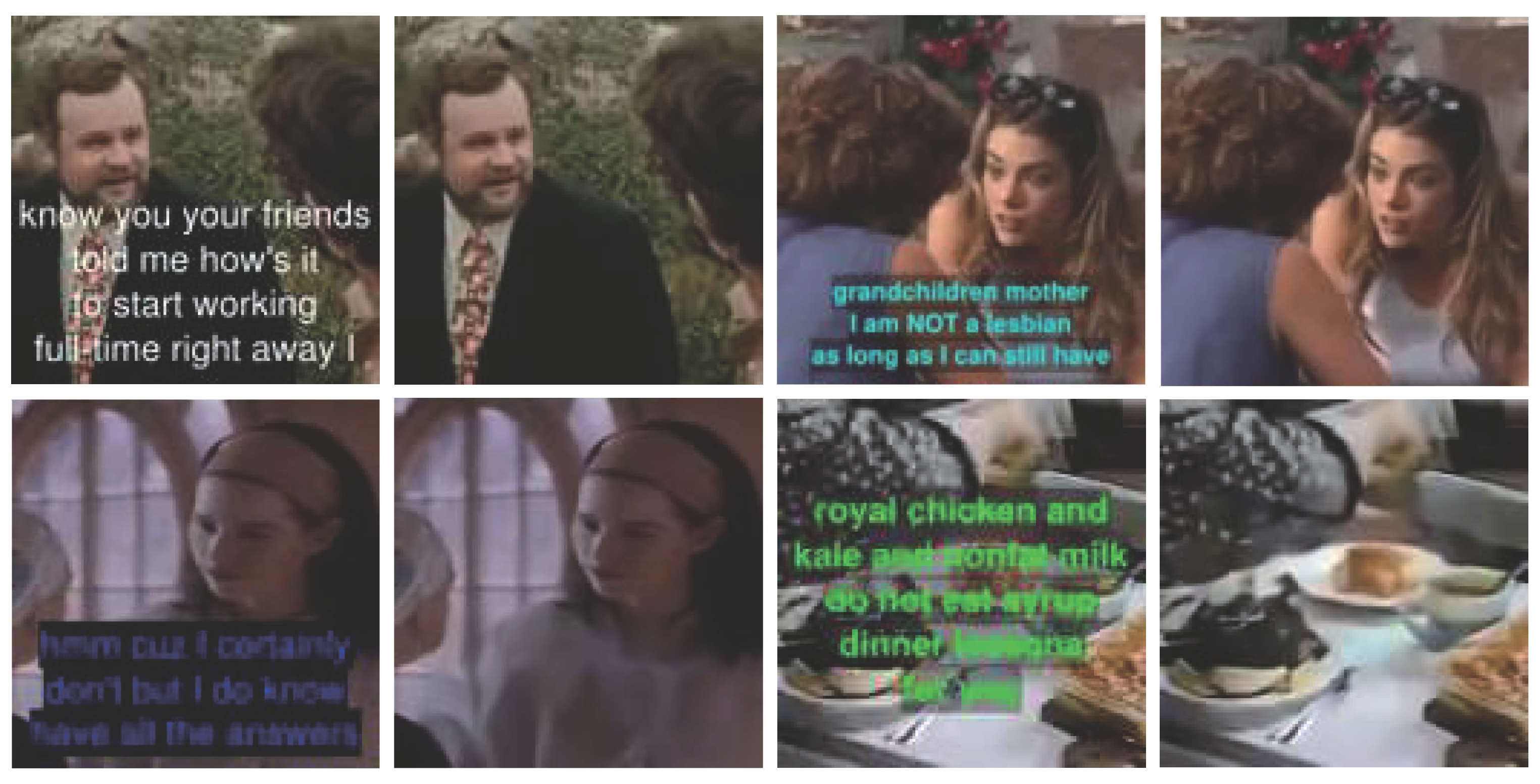}
\caption{Several representative examples of blind video decaptioning produced by~\citep{chu2021decaptioning}. }
\label{fig:video_decaptioning}
\end{center}
\end{figure}

\subsection{Applications}

\subsubsection{Blind Video Decaptioning} 
Blind video decaptioning aims to automatically remove subscripts and recover the occluded regions in videos without mask information. \cite{kim2019decaptioning} designed an encoder-decoder framework based on 3D convolution. They applied residual learning to directly touch the corrupted regions and leveraged feedback connections to enforce temporal coherence with the warping loss. However, this method often suffers from the problem of incomplete subtitle removal. \cite{chu2021decaptioning} proposed a two-stage video decaptioning network including a mask extraction module and a frame attention-based decaptioning module. Several examples produced by~\citep{chu2021decaptioning} are shown in Fig.~\ref{fig:video_decaptioning}. The regions originally covered by subtitles are filled with plausible content.

\subsubsection{Dynamic Object Removal} 
\begin{figure}[t]
\begin{center}
\includegraphics[width=\linewidth]{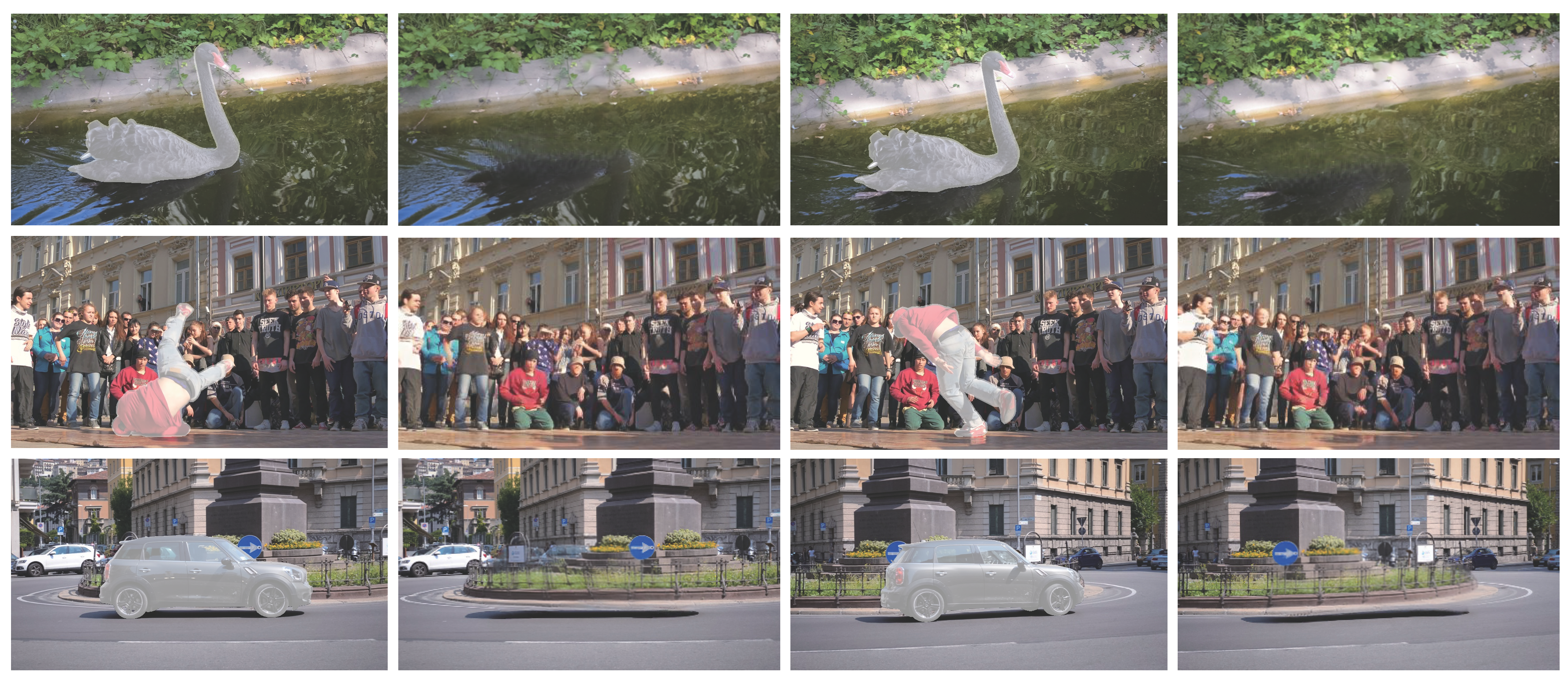}
\caption{Three examples of dynamic object removal produced by FuseFormer~\citep{liu2021fuseformer}, DlFormer~\citep{ren2022dlformer}, and ECFVI~\citep{kang2022error}. }
\label{fig:dynamic_object_removal}
\end{center}
\end{figure}

A common practical application of video inpainting technology is to automatically remove undesired objects, which are static or dynamic at the time of recording. In this part, we show two examples of dynamic object removal with the recent video inpainting methods~\citep{liu2021fuseformer,ren2022dlformer,kang2022error}. As shown in Fig.~\ref{fig:dynamic_object_removal}, the regions covered by dynamic objects can be filled with plausible content.

\section{Future Work}
\label{sec: future work}
Image and video inpainting essentially is a conditional generative task, therefore, the common generative models, such as VAE and GAN, are often adopted by the existing inpainting methods. Currently, diffusion models have become the most popular generative models with powerful capability of content synthesis. DMs would have the potential to improve the performance of image and video inpainting and may attract a lot of research effort in the future. For this promising direction, several challenging problems need to be solved.

\textbf{How to use large pre-trained diffusion models (\eg, denoising diffusion) for image inpainting?} \\
DMs synthesize an image by a sequential application of denoising steps, which are conducted in pixel or latent space. For the inpainting task, the core idea is to fill in the missing regions while preserving the originally valid content. Some researchers have made preliminary attempts, such as Palette~\citep{Palette2022}, Blended Diffusion~\citep{Avrahami2022Blended}, and ControlNet~\citep{zhang2023adding}, etc. One research challenge is how to inject conditioning information into the denoising processes of large pre-trained diffusion models. Following the pipeline of diffusion models, they need many iterations to generate the final image and thus require a longer inference time compared to existing VAE- and GAN-based approaches. Another research challenge is to implement fast inpainting methods based on diffusion. Also, while video-based generative diffusion models are still in their infancy, it is expected that large pre-trained video generation models will become available in the near future. Leveraging these models for video inpainting will be an interesting task once these models become available.

\textbf{How to use large pre-trained models for joint text and image embedding (\eg, the latest CLIP style architecture) for image inpainting?} \\
Mainstream inpainting methods are uncontrollable, where the inpainted content is unknown in advance and sometimes this is undesired for users. Reference-based inpainting cannot fully satisfy this requirement. On the other hand, recent studies~\citep{Rombach2022high,hertz2022prompt,Parmar2023zero} have shown that large pre-trained diffusion models with massive text-image pairs can synthesize high-quality images with rich low-level attributes and details. In addition, ~\cite{zhao2023unleashing} implied that such pre-trained DMs also contain high-level visual concepts. As a result, text-guided inpainting based on the large pre-trained text-to-image diffusion models would be able to fill the content under the control of users. The first problem is to design the appropriate prompt exactly indicating the user's intention. It is also challenging to merge the image embedding from the user prompt with the corresponding embedding of the input corrupted image. In addition, text-based video inpainting will be a great avenue for future work.

\textbf{How to scale up training to datasets of 5B images (e.g. LAION)?}\\
Deep learning models are hungry for training datasets. Currently, advanced diffusion models are pre-trained on large-scale datasets containing millions or even billions of text-image data pairs. However, these models are mainly dominated by several industrial research labs, where the datasets and training processes are not transparent to the research community. Very recently, the largest text-image dataset LAION-5B~\citep{schuhmann2022laionb} containing ~5.8 billion samples is publicly available. In future work, it is worth designing efficient methods for image and video inpainting that are trained on such very large datasets directly.

\textbf{How to utilize image data and pre-trained image inpainting models to improve the models of video inpainting?} In addition to considering the spatial aspect as in image inpainting, video inpainting also needs to consider the temporal aspect. Therefore, it is important and beneficial to transfer the inpainting ability from image to video. A simple and direct solution is to take the result of image inpainting on each frame as the initialization and then revise the spatial and temporal consistency via carefully designed deep models. Another possible research line is to take the well-trained image inpainting models as the backbone and aggregate the multiple frames in the feature space with appropriate modules, such as deformable convolution or attention. 
It's still worth exploring combining the pre-trained image inpainting model with deep video prior. 

\textbf{How to create a large video dataset of 5B videos and leverage it for video inpainting?} \\
Like image inpainting, taking advantage of large pre-trained text-video diffusion models may be a new research direction for video inpainting. However, current text-video DMs are trained on datasets with ~10 million captioned videos, which inevitably limits the generation and generalization ability of DMs. One potential direction of future research is to collect large-scale text-video datasets (\eg, 5B pairs) and design the pre-training methods scaling up to this amount. As we all know, video inpainting is more difficult compared to its image counterpart. Therefore, it is valuable to spend time on all aspects of large video datasets: building large publicly available video datasets, generating large diffusion methods for video synthesis and using these pre-trained methods for video inpainting, and separately designing and training large-scale video architectures directly.

\section{Concluding Remarks}
\label{sec:conclusions}

The prevalence of visual data, including images and video, promotes the development of related processing technologies, \eg, image and video inpainting. Due to their practical applications in many fields, these techniques have attracted great attention from both the industrial and research communities over the past decade. We presented a review of deep learning-based methods for image and video inpainting. Specifically, we outline different aspects of the research, including a taxonomy of existing methods, training objectives, benchmark datasets, evaluation protocols, performance evaluation, and real-world applications. Future research directions are also discussed.

Although current deep learning-based inpainting approaches have achieved remarkable performance improvement, there are still several limitations: (1) Uncertainty of artifacts. The results generated by inpainting methods often exhibit visual artifacts, which are difficult to predict and prevent. There is almost no research work to systematically and comprehensively study these artifacts. (2) Specificity. Current inpainting models are usually trained on specific datasets, for example, face images or natural scene images. In other words, models trained on face images have bad predictions on natural scene images, and vice versa. Not enough models are trained on large scale datasets such as LIAON. (3) Large-scale inpainting. Current advanced inpainting methods still have limited performance on large-scale missing regions. Many methods are based on attention mechanisms, which are more fragile in large-scale scenarios. (4) High training costs. Current deep learning-based inpainting methods often need one or more weeks on multiple GPUs, which places very high demands on resource consumption. (5) Long inference time. Diffusion model-based methods can achieve better inpainting performance, however, they need a very long running time, which limits the application scope of inpainting techniques. 

Deep image/video inpainting techniques have a wide range of real-world applications, however, they also raise potential ethical issues that need to be carefully considered and addressed: (1) Security risks. Inpainting-based visual data editing, \eg, object removal, may maliciously be exploited, such as tampering with visual data or altering evidence. (2) Ownership and copyright. When there is no appropriate authorization, deep inpainting techniques used to manipulate and enhance images/videos could raise questions about ownership and copyright. The inpainting result may strongly resemble or be strongly inspired by copyrighted material. (3) Historical accuracy. Inpainting methods can be used for the restoration of old photos/films or artworks. This process could raise risks of inadvertently changing the initial creative intention or historical accuracy of the content, which requires careful verification by domain experts. (4) Bias. If not properly trained, an inpainting model may introduce bias or unfairness, especially when the training data is biased or unrepresentative. This has the potential to perpetuate social prejudices or inaccurately portray certain groups. 


%
%

\bibliographystyle{spbasic}      
\bibliography{egbib}   

\end{document}